\documentclass[10pt,twocolumn,letterpaper]{article}
\usepackage[pagenumbers]{cvpr} 

\usepackage{graphicx}
\usepackage{amsmath}
\usepackage{amssymb}
\usepackage{booktabs}
\usepackage{enumitem}
\usepackage{cuted}
\usepackage{multirow}
\usepackage{xcolor}

\definecolor{cvprblue}{rgb}{0.21,0.49,0.74}
\usepackage[pagebackref,breaklinks,colorlinks,allcolors=cvprblue]{hyperref}

\usepackage[capitalize]{cleveref}
\crefname{section}{Sec.}{Secs.}
\Crefname{section}{Section}{Sections}
\Crefname{table}{Table}{Tables}
\crefname{table}{Tab.}{Tabs.}

\newcommand{\paravspace}{\vspace{-10pt}}

\title{Structured 3D Latents for Scalable and Versatile 3D Generation\thanks{Open-source project; see our \href{https://github.com/Microsoft/TRELLIS}{project page} for code, model, and data.}}

\makeatletter
\renewcommand*{\@fnsymbol}[1]{\ensuremath{\ifcase#1\or *\or \star\or \dagger\or \ddagger\or
		\mathsection\or \mathparagraph\or \|\or **\or \dagger\dagger
		\or \ddagger\ddagger \else\@ctrerr\fi}}
\makeatother

\author{
    Jianfeng Xiang$^{1,3}$\thanks{Work done during internship at Microsoft Research} \quad  Zelong Lv$^{2,3\star}$ \quad Sicheng Xu$^{3}$ \quad Yu Deng$^{3}$ \quad Ruicheng Wang$^{2,3\star}$\\
    Bowen Zhang$^{2,3\star}$\quad Dong Chen$^{3}$ \quad Xin Tong$^{3}$ \quad Jiaolong Yang$^{3}$\thanks{Corresponding author} \\
	$^1${Tsinghua University} \quad $^2${USTC} \quad $^3${Microsoft Research}\\
    \url{https://github.com/Microsoft/TRELLIS}
}

\begin{document}
\maketitle

\begin{strip}
    \vspace{-42pt}
	\centering
	\includegraphics[width=1\textwidth]{figures/compressed/teaser_v7.pdf}
    \vspace{-8pt}
    \captionsetup{type=figure,font=small,position=top}
    \caption{
    High-quality 3D assets generated by our method in various formats from text or image prompts (using GPT-4o and DALL-E 3). Our method enables versatile generation in about 10 seconds, offering vivid appearances with 3D Gaussians or Radiance Fields and detailed geometries with meshes. It also supports flexible 3D editing. \textbf{\emph{Best viewed with zoom-in.}}} 
    \label{fig:teaser}
    \vspace{-8pt}
\end{strip}

\begin{abstract}
We introduce a novel 3D generation method for versatile and high-quality 3D asset creation.
The cornerstone is a unified Structured LATent (\textsc{SLat}) representation which allows decoding to different output formats, such as Radiance Fields, 3D Gaussians, and meshes. This is achieved by integrating a sparsely-populated 3D grid with dense multiview visual features extracted from a powerful vision foundation model, comprehensively capturing both structural (geometry) and textural (appearance) information while maintaining flexibility during decoding.

We employ rectified flow transformers tailored for \textsc{SLat} as our 3D generation models and train models with up to 2 billion parameters on a large 3D asset dataset of 500K diverse objects. Our model generates high-quality results with text or image conditions, significantly surpassing existing methods, including recent ones at similar scales. We showcase flexible output format selection and local 3D editing capabilities which were not offered by previous models. 
\end{abstract}

\vspace{-20px}
\section{Introduction}\label{sec:introduction}

While AI Generated Content (AIGC) for 3D has made tremendous progress in recent years~\cite{poole2023dreamfusion,liu2023zero,tochilkin2024triposr}, existing 3D generative models still fall short in generation quality compared to their 2D predecessors, where large image generation models~\cite{esser2024scaling,chen2024pixart} have enabled ready-to-use tools that exert a profound impact on today's digital industry.

Unlike 2D images, typically represented by pixel grids, 3D data encompasses diverse representations like meshes, point clouds, Radiance Fields~\cite{mildenhall2021nerf}, and 3D Gaussians~\cite{kerbl20233d}. Each format is tailored for specific applications and may encounter difficulties when adapted for other tasks. For instance, while numerous studies~\cite{cheng2023sdfusion,gupta20233dgen,li2023generalized,ren2024xcube,xiong2024octfusion,zhang20233dshape2vecset,zhang2024clay} have utilized 3D representations like meshes or implicit fields~\cite{mescheder2019occupancy,park2019deepsdf} for object geometry generation, they often falter in detailed appearance modeling compared to those relying on representations equipped with advanced volumetric rendering capabilities (\eg, 3D Gaussians and Radiance Fields). Conversely, generative models based on Radiance Fields or 3D Gaussians~\cite{wang2023rodin,lan2024ln3diff,zhang2024gaussiancube} excel in rendering high-quality appearances but strruggle with plausible geometry extraction. Moreover, the unique structured or unstructured characteristics of different representations complicate processing through a consistent network architecture. These issues hinder the development of a standardized 3D generative modeling paradigm, in contrast to the consensus in recent advanced 2D generation methods that learn generative models within a unified latent space~\cite{rombach2022high,esser2024scaling}. 

In this paper, we aim to develop a \emph{unified and versatile latent space} that facilitates high-quality 3D generation across various representations, accommodating diverse downstream requirements. This problem is highly challenging and has rarely been  addressed by previous approaches. To tackle this, our primary strategy is to introduce explicit sparse 3D structures in the latent space design. These structures enable decoding into different 3D representations by characterizing attributes within the local voxels surrounding an object, as is evidenced by recent advancements in the 3D reconstruction field~\cite{lu2024scaffold,shen2023flexicubes,gao2023strivec}. This approach also allows for efficient high-resolution modeling by bypassing voxels without 3D information~\cite{liu2020neural,ren2024xcube}, and introduces locality that facilitates flexible editing.

However, even with such structures, achieving high-quality decoding into different 3D representations is still non-trivial, as it requires the latent representation to encapsulate both comprehensive geometry and appearance information of the 3D assets. To address this issue, our second strategy is to equip the sparse structures with a powerful vision foundation model~\cite{oquab2024dinov} for detailed information encoding, given its demonstrated strong 3D awareness~\cite{el2024probing} and capability for detailed representation~\cite{zou2024triplane}. This approach bypasses the need for a dedicated 3D encoder, and eliminates the costly pre-fitting process of aligning 3D data with specific representations~\cite{wang2023rodin,zhang2024gaussiancube}.

Given these two strategies, we introduce Structured LATents (\textsc{SLat}), a unified 3D latent representation for high-quality, versatile 3D generation. \textsc{SLat} marries \emph{sparse structures} with powerful \emph{visual representations}. It defines local latents on active voxels intersecting the object's surface. The local latents are encoded by fusing and processing image features from densely rendered views of the 3D asset, while attaches them onto active voxels. These features, derived from powerful pretrained vision encoders~\cite{oquab2024dinov}, capture detailed geometric and visual characteristics, complementing the coarse structure provided by the active voxels. Different decoders can then be applied to map \textsc{SLat} to diverse 3D representations of high quality. 

Building on \textsc{SLat}, we train a family of large 3D generation models, dubbed \emph{\textsc{Trellis}} in this paper, with text prompts or images as conditions. A two stage pipeline is applied which first generates the sparse structure of \textsc{SLat}, followed by generating the latent vectors for non-empty cells. We employ rectified flow transformers as our backbone models and adapt them properly to handle the sparsity in  \textsc{SLat}.
We train \textsc{Trellis} with up to 2 billion parameters on a large dataset of carefully-collected 3D assets.  Through extensive experiments, we show that our model can create high-quality 3D assets with detailed geometry and vivid texture, significantly surpassing previous methods. Moreover, it can easily generate 3D assets with different output formats to meet diverse downstream requirements. 

We summarize the notable features of our method below:
\begin{itemize}[leftmargin=2em]
	\item \textbf{High quality.} It produces diverse 3D assets at high-quality with intricate shape and texture details.\vspace{3pt}
	\item \textbf{Versatile generation.} It takes text or image prompts and can generate various final 3D representations including but not limited to Radiance Fields, 3D Gaussians, and meshes.\vspace{3pt}
	\item \textbf{Flexible editing.} It enables flexible tuning-free 3D editing such as the deletion, addition, and replacement of local regions, guided by text or image prompts.\vspace{3pt}
	\item \textbf{Fitting-free training.} No 3D fitting is needed for the training objects in the entire process.\vspace{3pt}
\end{itemize}
Given these strong performance and multifold advantages, we believe our new models can serve as powerful 3D generation foundations and unlock new possibilities for the 3D vision community.
We hope our work can shed some light on 3D-representation-agnostic asset modeling, in contrast to the field's relentless pursuit of and adaptation to new representations. 
\emph{All our code, model, and data are released to facilitate reproduction and downstream applications}.

\section{Related Works}\label{sec:related_works}

\paragraph{3D generative models.}
Early 3D generation methods primarily leveraged Generative Adversarial Nets (GANs)~\cite{goodfellow2014generative} to model 3D distributions~\cite{wu2016learning,zhu2018visual,deng2021gram,chan2022efficient,gao2022get3d,skorokhodov20233d,zheng2022sdf}, but faced challenges in scaling to more diverse scenarios.
Later approaches employed diffusion models~\cite{sohl2015deep,ho2020denoising} for various representations like point clouds~\cite{nichol2022point,luo2021diffusion}, voxel grids~\cite{tang2023volumediffusion,muller2023diffrf, hui2022neural}, Triplanes~\cite{chen2023single, shue20233d, wang2023rodin, zhang2024rodinhd}, and 3D Gaussians~\cite{zhang2024gaussiancube,he2024gvgen}. Some alternatives~\cite{nash2020polygen,chen2024meshanything} adopted GPT-style autoregressive models~\cite{radford2019language} for mesh generation. Despite these advancements, efficiency remains a challenge for generative modeling in raw data space. 

To enhance both quality and efficiency, recent studies have resorted to generation in a more compact latent space~\cite{rombach2022high}. Some methods~\cite{zhao2024michelangelo,vahdat2022lion, zhang20233dshape2vecset, zhang2024clay, li2024craftsman, wu2024direct3d, zheng2023locally, ren2024xcube} mainly focused on shape modeling, often requiring an additional texturing phase for complete 3D asset generation. 
Among them, a few approaches~\cite{gupta20233dgen, xiong2024octfusion} incorporated appearance information, but faced difficulties to model highly detailed appearance due to their surface representations. Other works~\cite{jun2023shap, lan2024ln3diff, ntavelis2023autodecoding, yang2024atlas} built latent representations for Radiance Fields or 3D Gaussians, which may pose challenges for accurate surface modeling. \cite{chen20243dtopia} encoded both geometry and appearance using latent primitives, but its pre-fitting process is both costly and lossy. 
In this work, we aim to build a versatile latent space that supports decoding into various 3D representations of high quality.

\paravspace
\paragraph{3D creation with 2D generative models.}

Instead of directly training 3D generative models, some recent methods leveraged 2D generative models to create 3D assets due to their superior generalization abilities. 
A pivotal work, DreamFusion~\cite{poole2023dreamfusion}, optimized 3D assets by distilling from pre-trained image diffusion models~\cite{rombach2022high}, followed by a large group of successors~\cite{tang2024dreamgaussian,lin2023magic3d,wang2024prolificdreamer,tang2023make,liang2024luciddreamer} with more advanced distillation techniques. Another group of works~\cite{liu2023zero,hong2024lrm,liu2024one,li2024instant3d,long2024wonder3d,shi2024mvdream,tang2024lgm,gslrm2024,zou2024triplane,xiang20233d,liu2024meshformer} involves generating multiview images via 2D diffusions and reconstructing 3D assets from them. However, these 2D-assisted approaches often yield lower geometry quality compared to native 3D models learned from 3D data collections, due to inherent multiview inconsistency in 2D generative models.

\paravspace
\paragraph{Rectified flow models.}
Rectified flow models~\cite{liu2023flow,albergo2023building,lipman2023flow} have recently emerged as a novel generative paradigm that challenges the dominance of diffusions~\cite{sohl2015deep,ho2020denoising}. 
Recent works~\cite{esser2024scaling,meta2024moviegen} have demonstrated the effectiveness of them for large-scale image and video generation. In this paper, we also apply rectified flow models and demonstrate their abilities for 3D generation at scale.

\section{Methodology}\label{sec:methodology}

\begin{figure*}[t]
	\centering
	\includegraphics[width=\textwidth]{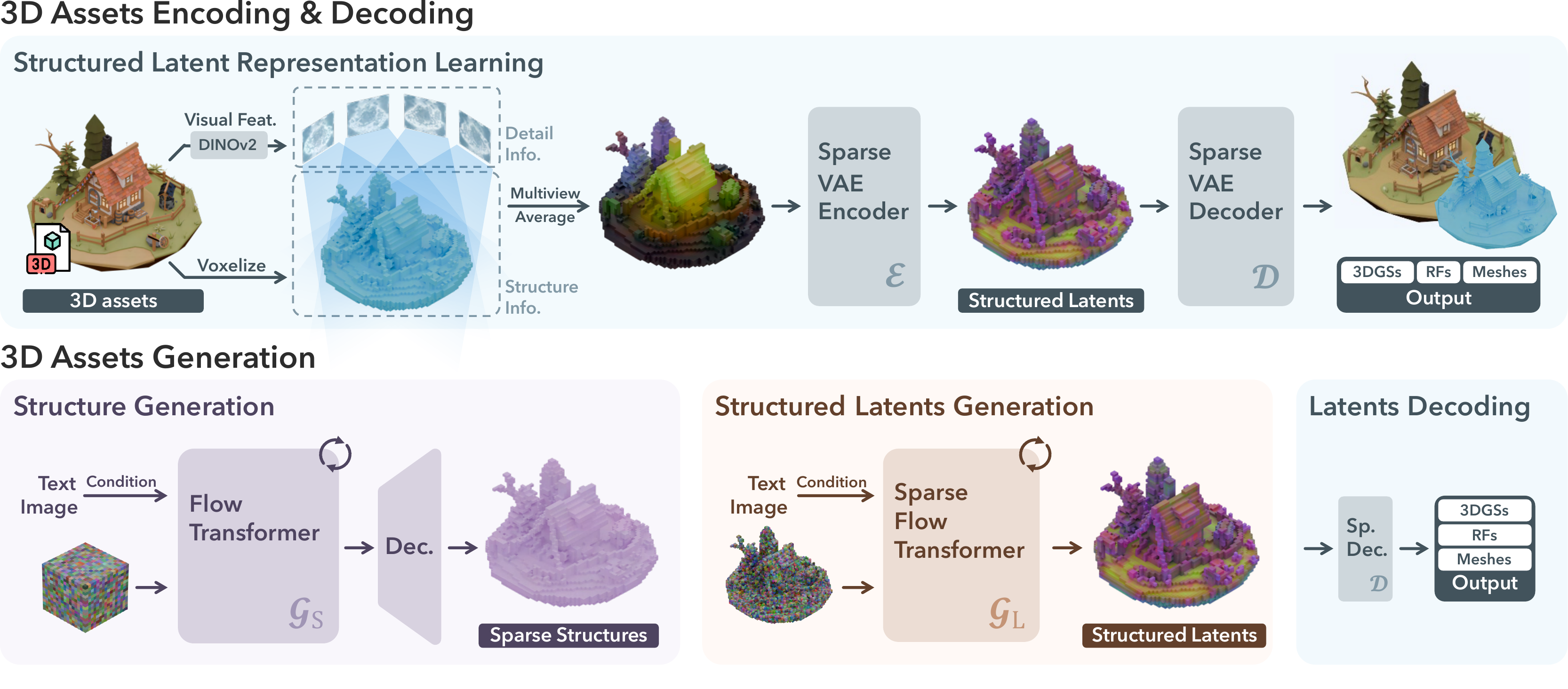}
    \vspace{-24pt}
	\caption{Overview of our method. \textbf{Encoding \& Decoding:} We adopt a structured latent representation (\textsc{SLat}) for 3D assets encoding, which defines local latents on a sparse 3D grid to represent both geometry and appearance information. It is encoded from the 3D assets by fusing and processing dense multiview visual features extracted from a DINOv2 encoder, and can be decoded into versatile output representations with different decoders. \textbf{Generation:} Two specialized rectified flow transformers are utilized to generate \textsc{SLat}, one for the sparse structure and the other for local latents attached to it. }
	\label{fig:pipeline}
    \vspace{-8pt}
\end{figure*}

We aim to generate high-quality 3D assets in various 3D representation formats given text or image conditions. Figure~\ref{fig:pipeline} shows an overview, with details described below.

\subsection{Structured Latent Representation}\label{sec:slate}
For a 3D asset $\mathcal{O}$, we encode its geometry and appearance information using a unified structured latent representation $\boldsymbol{z}$, which defines a set of local latents on a 3D grid:
\begin{equation}
  \small \boldsymbol{z} = \{(\boldsymbol{z}_i,\boldsymbol{p}_i)\}_{i=1}^{L},\quad \boldsymbol{z}_i\in\mathbb{R}^C, \ \boldsymbol{p}_i\in \{0, 1,\ldots, N-1\} ^3, \label{eq:slate}
\end{equation}
where $\boldsymbol{p}_i$ is the positional index of an active voxel in the 3D grid intersecting with the surface of $\mathcal{O}$, $\boldsymbol{z}_i$ denotes a local latent attached to the corresponding voxel, the derivation of which will be described later, $N$ is the spatial length of the 3D grid, and $L$ is the total number of active voxels.
Intuitively, the active voxels ${\boldsymbol{p}_i}$ outline the coarse structure of the 3D asset, while the latents ${\boldsymbol{z}_i}$ capture finer details of appearance and shape.
Together, these structured latents encompass the entire surface of $\mathcal{O}$, effectively capturing both the overall form and intricate details.

Due to the sparsity of 3D data, the number of active voxels is significantly smaller than the total size of the grid, \ie, $L \ll N^3$, allowing to be constructed at a relatively high resolution. By default, we set $N=64$ which leads to an average value of $L=20$K.

\subsection{Structured Latents Encoding and Decoding}\label{sec:encoding}
With the structured latent representation, we develop an effective encoding scheme to encode 3D assets to it, and introduce different decoders for reconstruction across various 3D representations. The details are outlined below.

\paravspace
\paragraph{Visual feature aggregation.} We first convert each 3D asset $\mathcal{O}$ into a voxelized feature $\boldsymbol{f}=\{(\boldsymbol{f}_i,\boldsymbol{p}_i)\}_{i=1}^{L}$.
Here, $\boldsymbol{p}_i$ is the active voxels as defined in Eq.~\eqref{eq:slate}, and $\boldsymbol{f}_i$ is a visual feature recording detailed structure and appearance information of the local region. 

To derive $\boldsymbol{f}_i$ for each active voxel, we aggregate features extracted from dense multiview images of $\mathcal{O}$. We render images from randomly sampled camera views on a sphere and extract feature maps using a pre-trained DINOv2 encoder~\cite{oquab2024dinov}. Each voxel is projected onto the multiview feature maps to retrieve features at corresponding locations, and their average is used as $\boldsymbol{f}_i$, as shown in Fig.~\ref{fig:pipeline} (left-top).
We set $\boldsymbol{f}$ to match the resolution of the structured latents $\boldsymbol{z}$ (\ie, $64^3$). Empirically, this is sufficient to reconstruct the original 3D asset at high fidelity, thanks to the strong representation capabilities of DINOv2 features together with the coarse structure provided by the active voxels.  

\paravspace
\paragraph{Sparse VAE for structured latents.} With the voxelized feature $\boldsymbol{f}$, we introduce a transformer-based VAE architecture for 3D assets encoding. 

Specifically, an encoder $\boldsymbol{\mathcal{E}}$ first encodes $\boldsymbol{f}$ to structured latents $\boldsymbol{z}$, followed by a decoder $\boldsymbol{\mathcal{D}}$ that converts $\boldsymbol{z}$ into a 3D asset represented by certain 3D representation. Reconstruction losses are then applied between the decoded 3D assets and the ground truth to train the encoder and decoder in an end-to-end manner, along with a KL-penalty on $\boldsymbol{z_i}$ to encourage normal distribution regularization following~\cite{rombach2022high}.

\begin{figure}[t]
    \centering
    \begin{subfigure}[b]{\linewidth}
        \centering  
        \includegraphics[width=\linewidth]{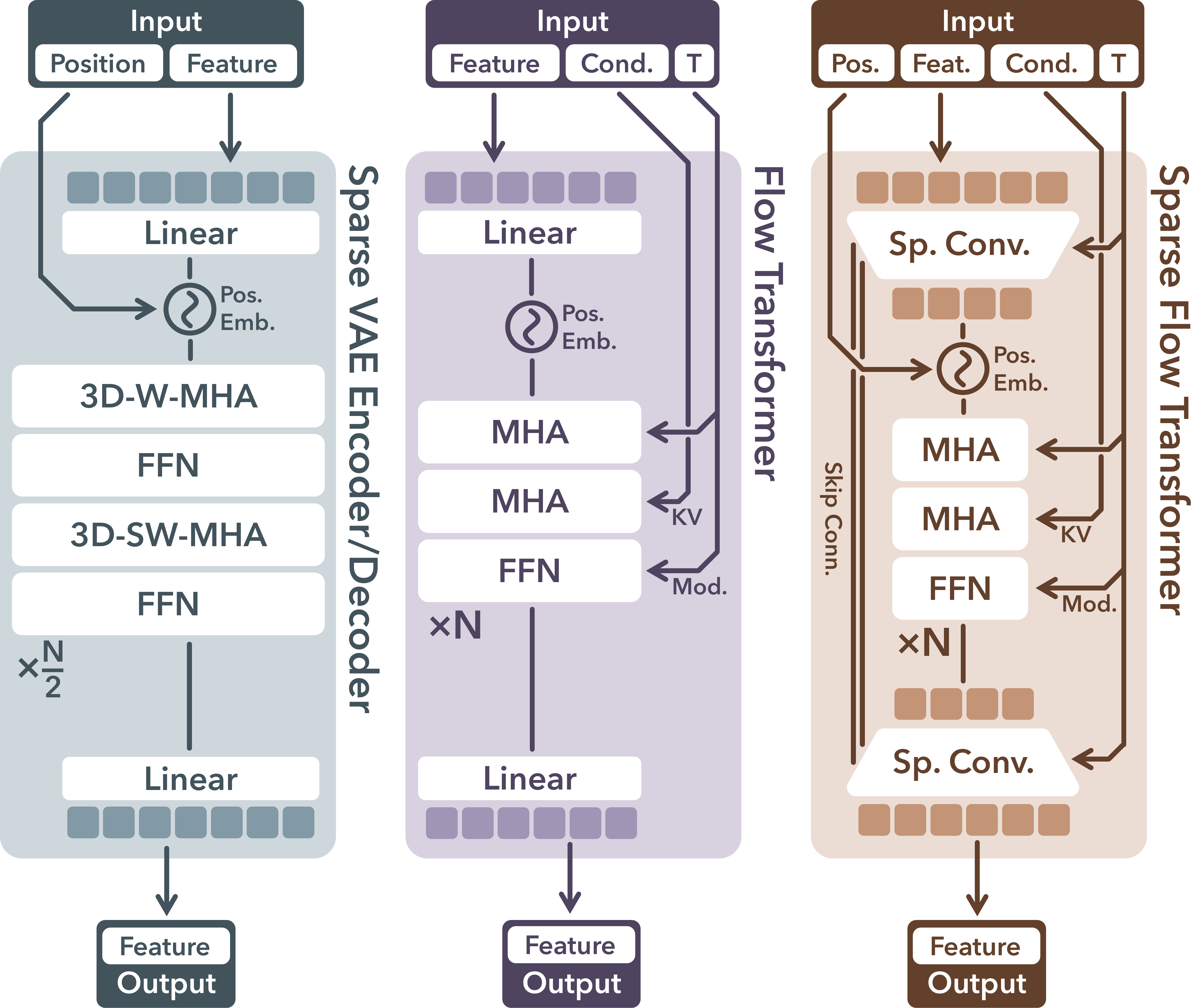}  
    \end{subfigure}

    \hspace{19pt}
    \begin{subfigure}[b]{0.1\linewidth}
        \caption{}  
        \label{fig:network_a}  
    \end{subfigure} 
    \hfill
    \begin{subfigure}[b]{0.1\linewidth}  
        \caption{}  
        \label{fig:network_b}  
    \end{subfigure} 
    \hfill
    \begin{subfigure}[b]{0.1\linewidth} 
        \caption{}
        \label{fig:network_c}  
    \end{subfigure} 
    \hspace{25pt}
    
    \vspace{-5pt}
    \caption{The network structures for encoding, decoding, and generation.}  
    \label{fig:network}  
    \vspace{-8pt}
\end{figure}

The encoder and decoder share the same transformer structure, as shown in Fig.~\ref{fig:network_a}. To handle sparse voxels, we serialize input features from active voxels and add sinusoidal positional encodings based on their voxel positions, creating tokens with variable context length $L$, which are subsequently processed through transformer blocks. Considering the locality characteristic of the latents, we incorporate shifted window attention~\cite{liu2021swin,yang2023swin3d} in 3D space to enhance local information interaction, which also improves efficiency compared to a full attention implementation. 

\paravspace
\paragraph{Decoding into versatile formats.} Our structured latents support decoding into diverse 3D representations, such as 3D Gaussians, Radiance Fields, and meshes, via respective decoders: $\boldsymbol{\mathcal{D}}_\mathrm{GS}$, $\boldsymbol{\mathcal{D}}_\mathrm{RF}$, and $\boldsymbol{\mathcal{D}}_\mathrm{M}$. These decoders share the same architecture except for their output layers, and can be trained using specific reconstruction losses tailored to their representations:

\vspace{4pt}
\noindent{\textit{(a) 3D Gaussians.}} The decoding process is formulated as:
\begin{equation}
    \boldsymbol{\mathcal{D}}_\mathrm{GS}\!:\{(\boldsymbol{z}_i,\boldsymbol{p}_i)\}_{i=1}^{L}\!\rightarrow\!\{\{(\small{\boldsymbol{o}_i^k,\boldsymbol{c}_i^k,\boldsymbol{s}_i^k,\alpha_i^k,\boldsymbol{r}_i^k})\}_{k=1}^{K}\}_{i=1}^{L},
\end{equation}
where each $\boldsymbol{z}_i$ is decoded into $K$ Gaussians with position offsets $\boldsymbol{o}$, colors $\boldsymbol{c}$, scales $\boldsymbol{s}$, opacities $\alpha$, and rotations $\boldsymbol{r}$. To maintain locality of $\boldsymbol{z_i}$, we constrain the final positions $\boldsymbol{x}$ of the Gaussians to the vicinity of their active voxel: $ \boldsymbol{x}^k_i = \boldsymbol{p}_i + \mathrm{tanh}(\boldsymbol{o}^k_i)$. The reconstruction losses consist of $\mathcal{L}_1$, D-SSIM and LPIPS~\cite{zhang2018unreasonable} between rendered Gaussians and the ground truth images. 

\vspace{3pt}
\noindent{\textit{(b) Radiance Fields.}} The decoding process is defined as:
\begin{equation}
    \boldsymbol{\mathcal{D}}_\mathrm{RF}\!:\{(\boldsymbol{z}_i,\boldsymbol{p}_i)\}_{i=1}^{L}\!\rightarrow\!\{(\boldsymbol{v}_i^\mathrm{x},\boldsymbol{v}_i^\mathrm{y},\boldsymbol{v}_i^\mathrm{z},\boldsymbol{v}_i^\mathrm{c})\}_{i=1}^{L},
\end{equation}
where $\boldsymbol{v}_i^\mathrm{x},\boldsymbol{v}_i^\mathrm{y},\boldsymbol{v}_i^\mathrm{z}\in\mathbb{R}^{16\times8}$ and $\boldsymbol{v}_i^\mathrm{c}\in\mathbb{R}^{16\times4}$ are the CP-decomposition of a local radiance volume at $8^3$ following Strivec~\cite{gao2023strivec}, while the reconstruction losses are similar to those for Gaussians. 

\vspace{3pt}
\noindent{\textit{(c) Meshes.}} The decoding process is as follows:
\begin{equation}
    \boldsymbol{\mathcal{D}}_\mathrm{M}\!:\{(\boldsymbol{z}_i,\boldsymbol{p}_i)\}_{i=1}^{L}\!\rightarrow\!\{\{(\boldsymbol{w}_i^j, d_i^j)\}_{j=1}^{64}\}_{i=1}^{L},
\end{equation}
where $\boldsymbol{w}_i^j\in\mathbb{R}^{45}$ are the flexible parameters in FlexiCubes~\cite{shen2023flexicubes} and $d_i^j\in\mathbb{R}^{8}$ is signed distance values for the eight vertices of the corresponding voxel. We append two convolutional upsampling blocks after the transformer backbone to increase the final output resolution to $256^3$ (\ie, each $\boldsymbol{z_i}$ for a grid of $4^3$), extract meshes from 0-level isosurfaces,  and compute $\mathcal{L}_1$ between rendered depth (normal) maps and their ground truth as the reconstruction losses. 

\vspace{3pt}
In practice, we adopt Gaussians to learn the encoder and decoder end-to-end due to their high fidelity and efficiency. For other output formats, we simply freeze the learned encoder and train their decoders from scratches as described above. Despite trained with Gaussians, the learned structured latents can faithfully reconstruct other formats, demonstrating strong extensibility (See Tab.~\ref{tab:reconstruction}). We leave more implementation details in Sec.~\ref{sec:training_details}.

\subsection{Structured Latents Generation}\label{sec:generation} 
We introduce a two-stage generation pipeline to generate the structured latents, which first generates the sparse structure, followed by the local latents attached to it. For modeling the latent distribution, we employ rectified flow models~\cite{lipman2023flow}. We will first provide a brief introduction to these models before detailing our generation pipeline.

\paravspace
\paragraph{Rectified flow models.} Rectified flow models use a linear interpolation forward process, $\boldsymbol{x}(t)=(1-t)\boldsymbol{x}_0+t\boldsymbol{\epsilon}$, which interpolates between data samples $\boldsymbol{x}_0$ and noises $\boldsymbol{\epsilon}$ with a timestep $t$. 
The backward process is represented as a time-dependent vector field, $\boldsymbol{v}(\boldsymbol{x},t) = \nabla_t\boldsymbol{x}$, 
moving noisy samples toward the data distribution, and can be approximated with a neural network $\boldsymbol{v}_\theta$ by minimizing the conditional flow matching (CFM) objective~\cite{lipman2023flow}:
\vspace{-1pt}
\begin{equation}
    \mathcal{L}_{CFM}(\theta)=\mathbb{E}_{t,\boldsymbol{x}_0,\boldsymbol{\epsilon}}\|\boldsymbol{v}_\theta(\boldsymbol{x}, t)-(\boldsymbol{\epsilon}-\boldsymbol{x}_0)\|^2_2. \label{eq:cfm}
\end{equation}
\vspace{-10pt}

\paravspace
\paragraph{Sparse structure generation.}

In the first stage, we aim to generate the sparse structure $\{\boldsymbol{p}_i\}_{i=1}^{L}$. To enable this with a tensorized neural network, we convert the sparse active voxels into a dense binary 3D grid $\boldsymbol{O} \in \{0,1\}^{N\times N\times N}$, setting voxel values to $1$ if active, and $0$ otherwise.

Directly generating the dense grid $\boldsymbol{O}$ is computationally expensive. We introduce a simple VAE with 3D convolutional blocks to compress it into a low-resolution feature grid $\boldsymbol{S}\in\mathbb{R}^{D\times D\times D \times C_\mathrm{S}}$. Since $\boldsymbol{O}$ represents only coarse geometry, this compression is nearly lossless, enhancing efficiency significantly. It also converts the discrete values in $\boldsymbol{O}$ into continuous features suited for rectified flow training.

We introduce a simple transformer backbone $\boldsymbol{\mathcal{G}}_{\mathrm{S}}$ for generating $\boldsymbol{S}$, as shown in Fig.~\ref{fig:network_b}. An input dense noisy grid is serialized, combined with positional encodings (as in Sec.~\ref{sec:encoding}),
and fed into the transformer for denoising. Timestep information is incorporated using adaptive layer normalization (adaLN) and a gating mechanism~
\cite{peebles2023scalable}. Conditions are injected through cross attention layers as keys and values. For text conditions, we use features from a pretrained CLIP~\cite{radford2021learning} model. For image conditions, we adopt visual features from DINOv2. The denoised feature grid $\boldsymbol{S}$ is decoded into the discrete grid $\boldsymbol{O}$, and further converted back to active voxels $\{\boldsymbol{p}_i\}_{i=1}^{L}$ as the final sparse structure. 

\paravspace
\paragraph{Structured latents generation.}
In the second stage, we generate latents $\{\boldsymbol{z}_i\}_{i=1}^{L}$ given the structure $\{\boldsymbol{p}_i\}_{i=1}^{L}$ using a transformer $\boldsymbol{\mathcal{G}}_{\mathrm{L}}$ designed for sparse structures (Fig.~\ref{fig:network_c}).

Instead of directly serializing input noisy latents as in the sparse VAE encoder in Sec.~\ref{sec:encoding}, we improve efficiency by packing them into a shorter sequence before serialization, similarly as done by DiT~\cite{peebles2023scalable}. Due to our sparse structure, we apply a downsampling block with sparse convolutions~\cite{wang2017ocnn} to pack latents within a $2^3$ local region, followed by multiple time-modulated transformer blocks. A convolutional upsampling block is appended at the end of the transformer, with skip connections to the downsampling block that facilitates spatial information flow. Like in $\boldsymbol{\mathcal{G}}_{\mathrm{S}}$, timesteps are integrated via adaLN layers, and text/image conditions are injected through cross-attentions.

We train $\boldsymbol{\mathcal{G}}_{\mathrm{S}}$ and $\boldsymbol{\mathcal{G}}_{\mathrm{L}}$ separately using the CFM objective in Eq.~\eqref{eq:cfm}. After training, structured latents $\boldsymbol{z}=\{(\boldsymbol{z}_i,\boldsymbol{p}_i)\}_{i=1}^{L}$ can be sequentially generated by the two models and converted into high-quality 3D assets in various formats by different decoders: $\boldsymbol{\mathcal{D}}_{\mathrm{GS}}$, $\boldsymbol{\mathcal{D}}_{\mathrm{RF}}$, and $\boldsymbol{\mathcal{D}}_{\mathrm{M}}$. See Sec.~\ref{sec:implement} for more details.

\subsection{3D Editing with Structured Latents}\label{sec:editing} 

Our method supports flexible 3D editing and we present two simple \emph{tuning-free} editing strategies.

\paravspace
\paragraph{Detail variation.} The separation between the structure and latents enables detail variation of 3D assets without affecting the overall coarse geometry. This can be easily accomplished by preserving the asset's structure and executing the second generation stage with different text prompts.

\paravspace
\paragraph{Region-specific editing.} The locality of \textsc{SLat} allows for region-specific editing by altering voxels and latents in targeted areas while leaving others unchanged. To this end, we adapt Repaint~\cite{lugmayr2022repaint} to our two-stage generation pipeline. Given a bounding box for the voxels to be edited, we modify our flow models' sampling processes to create new content in that region, conditioned on the unchanged areas and any provided text or image prompts. Consequently, the first stage generates new structures within the specified region, and the second stage produces coherent details. 

\begin{figure*}[ht]
	\centering
	\includegraphics[width=\linewidth]{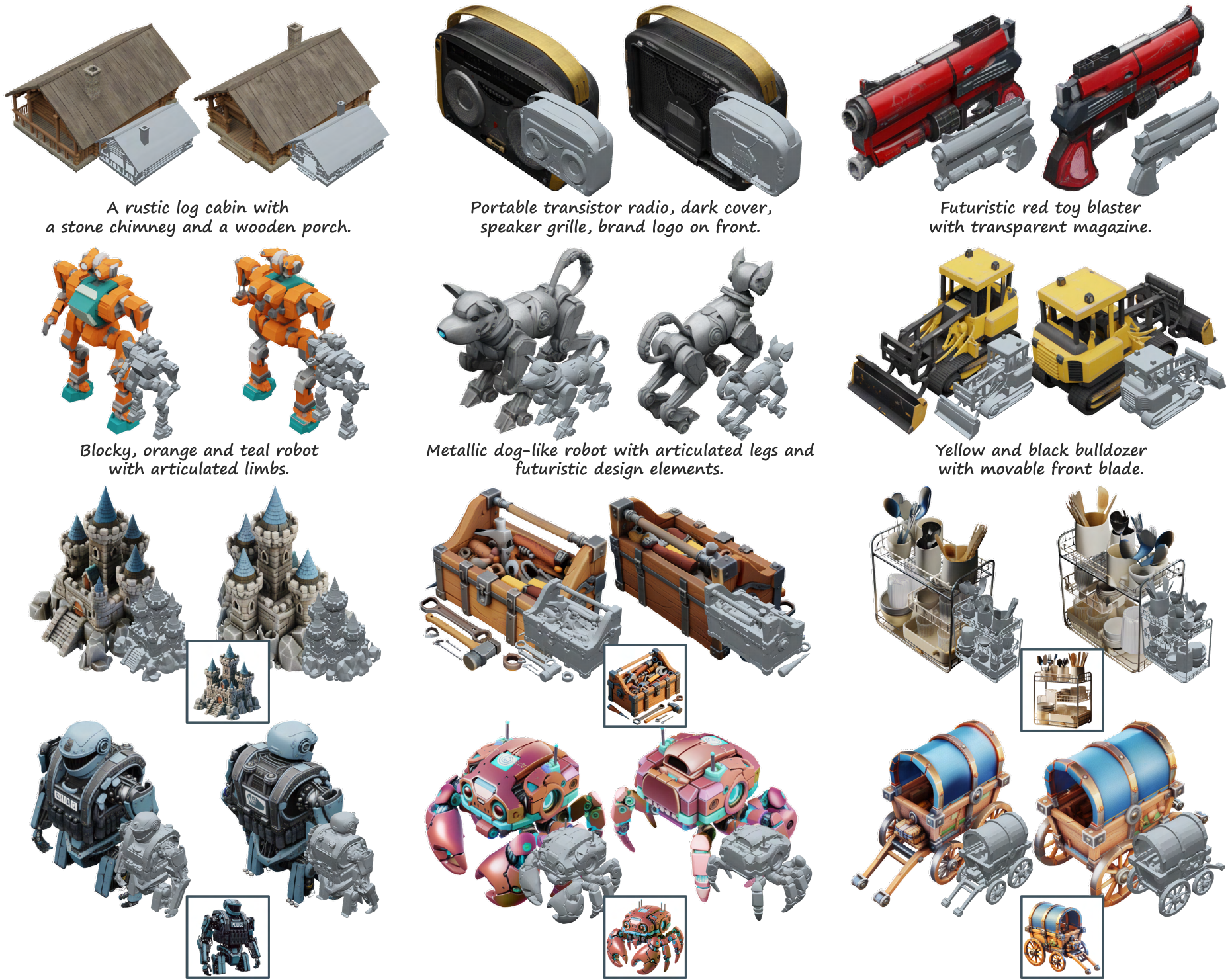}
    \vspace{-20pt}
	\caption{High-quality 3D assets created by our method, represented in Gaussians and meshes, given AI-generated text or image prompts.}
	\label{fig:results}
    \vspace{-8pt}
\end{figure*}

\section{Experiments}\label{sec:experiments}

\paragraph{Implementation details.}
For training, we carefully collect approximately 500K high-quality 3D assets from 4 public datasets: Objaverse (XL)~\cite{deitke2024objaverse}, ABO~\cite{collins2022abo}, 3D-FUTURE~\cite{fu20213d}, and HSSD~\cite{khanna2023hssd}. We render 150 images per asset, and employ GPT-4o~\cite{2024GPT4o} for captioning. 
Data augmentation is applied to both text and image prompts: texts are summarized to varying lengths, and images are rendered with different FoVs. We use classifier-free guidance (CFG)~\cite{ho2021classifier} with a drop rate of $0.1$ and AdamW~\cite{loshchilov2017decoupled} optimizer with a learning rate of $1e-4$. We train three models with total parameters of 342M (Basic), 1.1B (Large), and 2B (X-Large). The XL model is trained with 64 A100 GPUs (40G) for 400K steps with a batchsize of 256. At inference, CFG strength is set to $3$ and sampling steps to $50$. 

For quantitative evaluations, we use Toys4k~\cite{stojanov2021using}, which is not part of our training set or those of the compared methods. For visual results, comparisons, and user studies, we use text generated by GPT-4~\cite{achiam2023gpt} and images by DALL-E 3~\cite{betker2023improving}. Our method uses decoded \emph{Gaussians for appearance} evaluation and \emph{meshes for geometry}, unless specified otherwise. Refer to the \emph{suppl. material} for more details. 

\begin{table}[t]  
	\centering  
	\scriptsize
    \caption{Reconstruction fidelity of different latent representations. (\dag: evaluated using albedo color; \ddag: evaluated via Radiance Fields)}  
    \vspace{-8pt}
    \setlength{\tabcolsep}{2pt}
	\begin{tabular}{c|cc|cccc}  
        \toprule
        \multirow{2}{*}{\textbf{Method}} & \multicolumn{2}{c|}{\textbf{Appearance}} & \multicolumn{4}{c}{\textbf{Geometry}} \\
		 & \textbf{PSNR$\uparrow$} & \textbf{LPIPS$\downarrow$} & \textbf{CD$\downarrow$} & \textbf{F-score$\uparrow$} & \textbf{PSNR-N$\uparrow$} & \textbf{LPIPS-N$\downarrow$}  \\  
		 \midrule  
        LN3Diff & 26.44 & 0.076 & 0.0299 & 0.9649 & 27.10 & 0.094 \\
        3DTopia-XL & 
        25.34\textsuperscript{\dag} & 0.074\textsuperscript{\dag} & 0.0128 & 0.9939 & 31.87 & 0.080 \\
        CLAY & -- & -- & 0.0124 & 0.9976 & 35.35 & 0.035 \\
		\textbf{Ours} & \textbf{32.74}/\tiny{32.19\textsuperscript{\ddag}} & \textbf{0.025}/\tiny{0.029\textsuperscript{\ddag}} & \textbf{0.0083} & \textbf{0.9999} & \textbf{36.11} & \textbf{0.024} \\  
		\bottomrule
	\end{tabular}  
    \vspace{-8pt}
	\label{tab:reconstruction}  
\end{table} 

\begin{figure*}[t]
	\centering
	\includegraphics[width=1\linewidth]{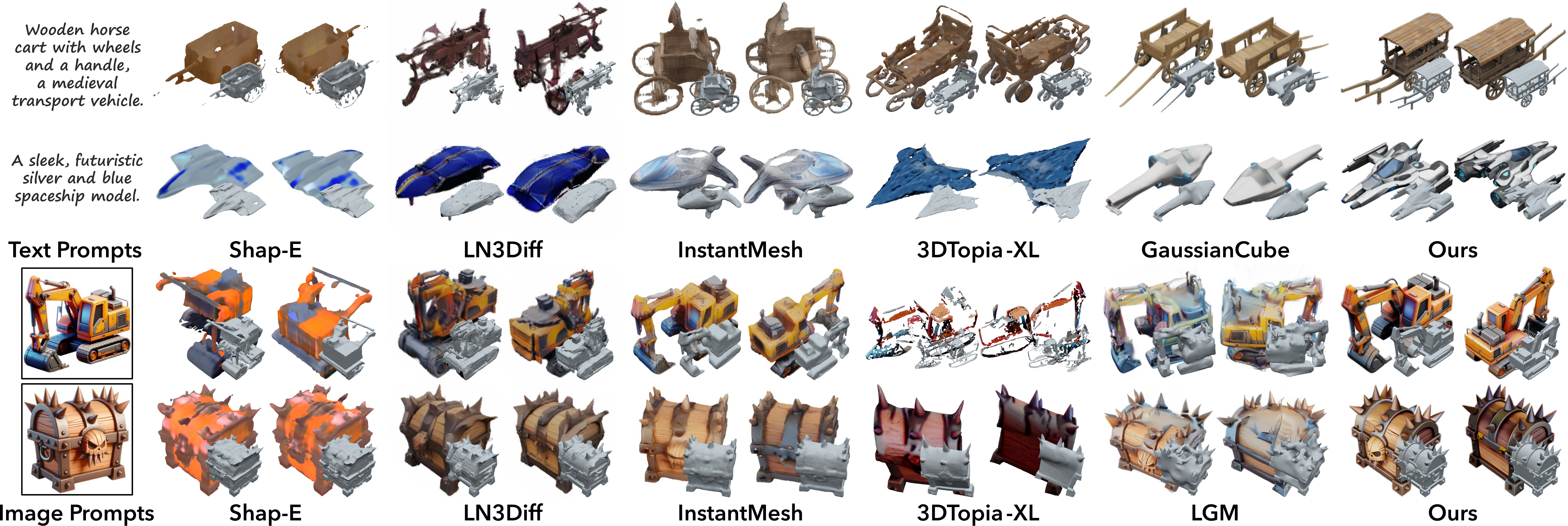}
    \vspace{-20pt}
	\caption{Visual comparisons of generated 3D assets between our method and previous approaches, given AI-generated prompts.}
	\label{fig:comparisons}
\end{figure*}

\begin{table*}[t]  
	\centering  
	\scriptsize
    \caption{Quantitative comparisons using Toys4k~\cite{stojanov2021using}. (KD is reported $\times100$. \dag: evaluated using shaded images of PBR meshes.)}
    \vspace{-8pt}
	\setlength{\tabcolsep}{4pt}
	\begin{tabular}{c|cccccc|cccccc}  
		\toprule 
		\multirow{2}{*}{\textbf{Method}} & \multicolumn{6}{c|}{\textbf{Text-to-3D}} & \multicolumn{6}{c}{\textbf{Image-to-3D}} \\
        & $\textbf{CLIP}\!\uparrow$ & $\textbf{FD}_\textbf{incep}\!\downarrow$ & $\textbf{KD}_\textbf{incep}\!\downarrow$ & $\textbf{FD}_\textbf{dinov2}\!\downarrow$ & $\textbf{KD}_\textbf{dinov2}\!\downarrow$ &$\textbf{FD}_\textbf{point}\!\downarrow$ & $\textbf{CLIP}\!\uparrow$ & $\textbf{FD}_\textbf{incep}\!\downarrow$ & $\textbf{KD}_\textbf{incep}\!\downarrow$ & $\textbf{FD}_\textbf{dinov2}\!\downarrow$ & $\textbf{KD}_\textbf{dinov2}\!\downarrow$ &$\textbf{FD}_\textbf{point}\!\downarrow$ \\
		\midrule
		Shap-E & 25.04 & 37.93 & 0.78 & 497.17 & 49.96 & 6.58 & 82.11 & 34.72 & 0.87 & 465.74 & 62.72 & 8.20 \\
		LGM &  24.83 & 36.18 & 0.77 & 507.47 & 61.89 & 24.73 & 83.97 & 26.31 & 0.48 & 322.71 & 38.27 & 15.90  \\ 
        InstantMesh & 25.56 & 36.73 & 0.62 & 478.92 & 49.77 & 10.79 & 84.43 & 20.22 & 0.30 & 264.36 & 25.99 & 9.63 \\ 
		3DTopia-XL & 22.48\textsuperscript{\dag} & 53.46\textsuperscript{\dag}
 & 1.39\textsuperscript{\dag} & 756.37\textsuperscript{\dag} & 87.40\textsuperscript{\dag} & 13.72 & 78.45\textsuperscript{\dag} & 37.68\textsuperscript{\dag} & 1.20\textsuperscript{\dag} & 437.37\textsuperscript{\dag} & 53.24\textsuperscript{\dag} &	18.21 \\ 
        Ln3Diff & 18.69 & 71.79 & 2.85 & 976.40 & 154.18 & 19.40 & 82.74 & 26.61 & 0.68 & 357.93 & 50.72 & 7.86 \\
        GaussianCube & 24.91 & 27.35 & 0.30 & 460.07 & 39.01 & 29.95 & -- & -- & -- & -- & -- & -- \\ 
		\textbf{Ours L} & \underline{26.60} & \underline{20.54} & \textbf{0.08} & \underline{238.60} & \underline{4.24} & \underline{5.24} & \textbf{85.77} & \textbf{9.35} & \textbf{0.02} & \textbf{67.21} & \textbf{0.72} & \textbf{2.03} \\  
		\textbf{Ours XL} & \textbf{26.70} & \textbf{20.48} & \textbf{0.08} & \textbf{237.48} & \textbf{4.10} & \textbf{5.21} & -- & -- & -- & -- & -- & -- \\  
		\bottomrule
	\end{tabular}  
	\vspace{-8pt}
	\label{tab:comparison}  
\end{table*} 

\subsection{Reconstruction Results}
\label{sec:recon}

We first assess the reconstruction fidelity of different latent representations. We compare \textsc{SLat} with alternatives also learned from large-scale data: latent point clouds from 3DTopia-XL~\cite{chen20243dtopia}, latent vector sets from CLAY~\cite{zhang2024clay}, and latent triplanes from LN3Diff~\cite{lan2024ln3diff}. 

For appearance fidelity, we report PSNR and LPIPS between rendered reconstruction results and ground truth. For geometry quality, we use Chamfer Distance (CD) and F-score to assess overall shape accuracy, and PSNR and LPIPS for rendered normal maps to evaluate surface details. 

As shown in Tab.~\ref{tab:reconstruction}, our method outperforms all baselines across all evaluated metrics. For geometry, it even surpasses CLAY which focuses solely on shape encoding. The high-fidelity reconstruction results under diverse output formats demonstrates strong versatility of \textsc{SLat}.

\subsection{Generation Results}\label{sec:gen_results}

In this section, we evaluate our generation quality. We first present various 3D generation results of our method, and then compare with other baseline methods.

\paravspace
\paragraph{Text/image-to-3D generation.}
Figure~\ref{fig:results} showcases 3D assets generated by our method, where the text and image prompts are given below. We present two views for each asset: front-left and back-right.

Upon visual inspection, our method produces 3D assets with an unprecedented level of quality. The generated appearances possess vibrant colors and vivid details, such as the radio speaker's grille and the toy blaster's scratches. The geometries reveal complex structures and fine shape details, with superior surface properties like flat faces and sharp edges (\eg, the bulldozer's hollow driving cab and the equipment on the police robot). It can even handle \emph{translucent objects} such as the 
drinking glasses on the kitchen rack.
Additionally, the generated contents closely match the elements from the provided text (\eg, the log cabin with a stone chimney and wooden porch) and faithfully adhere to details from input images (\eg, the castle with brick walls). More results can be found in Fig.~\ref{fig:teaser} and Sec.~\ref{sec:more_results}.

\begin{figure}[t]
	\centering
	\includegraphics[width=\linewidth]{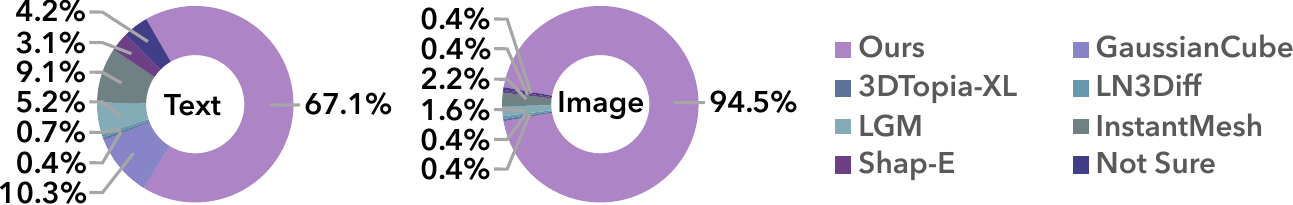}
    \vspace{-16pt}
	\caption{User study for text/image-to-3D generation.}
	\label{fig:user_study}
    \vspace{-8pt}
\end{figure}

\paravspace
\paragraph{Qualitative comparisons.}
We compare our approach with existing 3D generation methods that utilize different generative paradigms, latent representations, and output formats, including 2D-assisted methods: InstantMesh~\cite{xu2024instantmesh} and LGM~\cite{tang2024lgm}; and 3D generative approaches: 
GaussianCube~\cite{zhang2024gaussiancube}, Shap-E~\cite{jun2023shap}, 3DTopia-XL, and LN3Diff. We do not compare with CLAY in this phase, as their generation models are currently unavailable to us.

We begin by presenting visual comparisons in Fig.~\ref{fig:comparisons}. Our method outperforms all previous approaches, offering not only more vivid appearances and finer geometries but also more precise alignment with the provided text and image prompts.
It excels at producing intricate and coherent details, whereas alternatives experience varying degrees of quality degradation: The 2D-assisted methods suffer from structural distortion due to multiview inconsistencies inherent in the 2D generative models they rely on; other 3D generative approaches encounter featureless appearances and geometries, constrained by the limited reconstruction fidelity of their latent representations. GaussianCube and LGM do not provide plausible geometries, which is an inherent issue with their 3D Gaussian representations.

\paravspace
\paragraph{Quantitative comparisons.} Furthermore, we perform quantitative comparisons using text and image prompts in Toys4k and present the results in Tab.~\ref{tab:comparison}. We utlize Fréchet distance (FD)~\cite{heusel2017gans} and kernel distance (KD)~\cite{binkowski2018demystifying} with various feature extractors (\ie, Inception-v3~\cite{szegedy2016rethinking}, DINOv2, and PointNet++~\cite{qi2017pointnet++}) to assess overall quality of the generated outputs, and use CLIP score~\cite{radford2021learning} to evaluate the 
consistency between the generated results and the input prompts. As demonstrated, our method significantly surpasses previous methods across all evaluated metrics. 

\paravspace
\paragraph{User study.}
In addition, we conduct a user study with over 100 participants to compare different methods based on human preferences. We leverage 68 AI-generated text prompts and 67 image prompts, and create 3D assets from them via each method without any curation. As illustrated in Fig.~\ref{fig:user_study}, our method is strongly preferred by users due to its significant improvements in generation quality. Details of the user study can be found in Sec.~\ref{sec:user_study_detail}.

\begin{table}[t]  
	\centering  
	\scriptsize
    \caption{Ablation study on the size of \textsc{SLat}.}
    \vspace{-4pt}
	\setlength{\tabcolsep}{4pt}
    \begin{tabular}{cc|cc}  
		\toprule
        \textbf{Resolution} & \textbf{Channel} & \textbf{PSNR$\uparrow$} & \textbf{LPIPS$\downarrow$} \\
		\midrule
        32 & 16 & 31.64 & 0.0297 \\
        32 & 32 & 31.80 & 0.0289 \\
        32 & 64 & \underline{31.85} & \underline{0.0283} \\
        64 & 8 & \textbf{32.74} & \textbf{0.0250} \\
		\bottomrule  
	\end{tabular} 
    \vspace{-4pt}
	\label{tab:ablation_resolution}  
\end{table} 

\begin{table}[t]  
	\centering  
	\scriptsize
    \caption{Ablation study on different generation paradigms.}
    \vspace{-4pt}
	\setlength{\tabcolsep}{4pt}
	\begin{tabular}{cc|cc|cc}  
		\toprule
		& \multirow{2}{*}{\textbf{Method}} & \multicolumn{2}{c|}{\textbf{Training set}} & \multicolumn{2}{c}{\textbf{Toys4k}}\\
        && $\textbf{CLIP}\!\uparrow$ & $\textbf{FD}_\textbf{dinov2}\!\downarrow$ & $\textbf{CLIP}\!\uparrow$ & $\textbf{FD}_\textbf{dinov2}\!\downarrow$ \\  
		\midrule
        \multirow{2}{*}{Stage 1}
        & Diffusion & 25.09 & 132.71 &  25.86 &  295.90  \\
        & Rectified flow & \textbf{25.40} & \textbf{113.42} & \textbf{26.37} &  \textbf{269.56} \\
		\midrule
        \multirow{2}{*}{Stage 2}
        & Diffusion & 25.58 & 100.88 & 26.45 & 244.08 \\
        & Rectified flow & \textbf{25.65}  & \textbf{95.97} & \textbf{26.61} & \textbf{240.20} \\
		\bottomrule  
	\end{tabular} 
    \vspace{-4pt}
	\label{tab:ablation_frameworks}  
\end{table} 

\begin{table}[t]
    \centering  
    \scriptsize
    \caption{Ablation study on model size.}
    \vspace{-4pt}
    \setlength{\tabcolsep}{4pt}
    \begin{tabular}{c|cc|cc}  
        \toprule  
        \multirow{2}{*}{\textbf{Method}} & \multicolumn{2}{c|}{\textbf{Training set}} & \multicolumn{2}{c}{\textbf{Toys4k}}\\
        & $\textbf{CLIP}\!\uparrow$ & $\textbf{FD}_\textbf{dinov2}\!\downarrow$ & $\textbf{CLIP}\!\uparrow$ & $\textbf{FD}_\textbf{dinov2}\!\downarrow$ \\  
        \midrule
        B & 25.41 & 121.45 & 26.47 & 265.26 \\
        L & \underline{25.62} & \underline{99.92} & \underline{26.60} & \underline{238.60} \\
        XL & \textbf{25.71} & \textbf{93.96} & \textbf{26.70} & \textbf{237.48} \\
        \bottomrule  
    \end{tabular}  
    \label{tab:ablation_scale}  
 \end{table}

\subsection{Ablation Study}\label{sec:ablation}
We conduct ablation studies to validate the design choices of our method under the text-to-3D configuration. 

\paravspace
\paragraph{Size of structured latents.} To determine the size for \textsc{SLat}, we train sparse VAEs with varying latent resolutions and channels. As shown in Tab.~\ref{tab:ablation_resolution}, while the performance under $32^3$ is quite good, it tends to plateau as the number of latent channels increases. Switching to $64^3$ provides a significant boost. We prioritize quality over efficiency and adopt $64^3$ as our default setting for \textsc{SLat}.

\paravspace
\paragraph{Rectified flow \emph{v.s.} diffusion.}
We compare rectified flow models with a widely used diffusion baseline~\cite{peebles2023scalable} in Tab.~\ref{tab:ablation_frameworks}. We independently alter the generation method at each stage using the large model size, while maintaining the XL model unchanged for the other stages. As shown, replacing diffusion models with rectified flow models at any stage improves both generation quality and prompt alignment.

\paravspace
\paragraph{Model size.}
We examine the model's performance with varying numbers of parameters. Table~\ref{tab:ablation_scale} shows that increasing the model size consistently improves the generation performance on both training distribution and Toys4k.

\subsection{Applications}

We demonstrate tuning-free applications of our method by utilizing the editing strategies described in Sec.~\ref{sec:editing}.

\begin{figure}[t]
	\centering
    \begin{minipage}[b]{0.05\linewidth}
        \begin{subfigure}[b]{\linewidth}
            \subcaption{}
            \label{fig:asset_variation}
        \end{subfigure}
        \vspace{65pt}
        
        \begin{subfigure}[b]{\linewidth}
            \subcaption{}
            \label{fig:local_editing}
        \end{subfigure}
        \vspace{17pt}
    \end{minipage} 
    \begin{minipage}[b]{0.9\linewidth}
        \includegraphics[width=\linewidth]{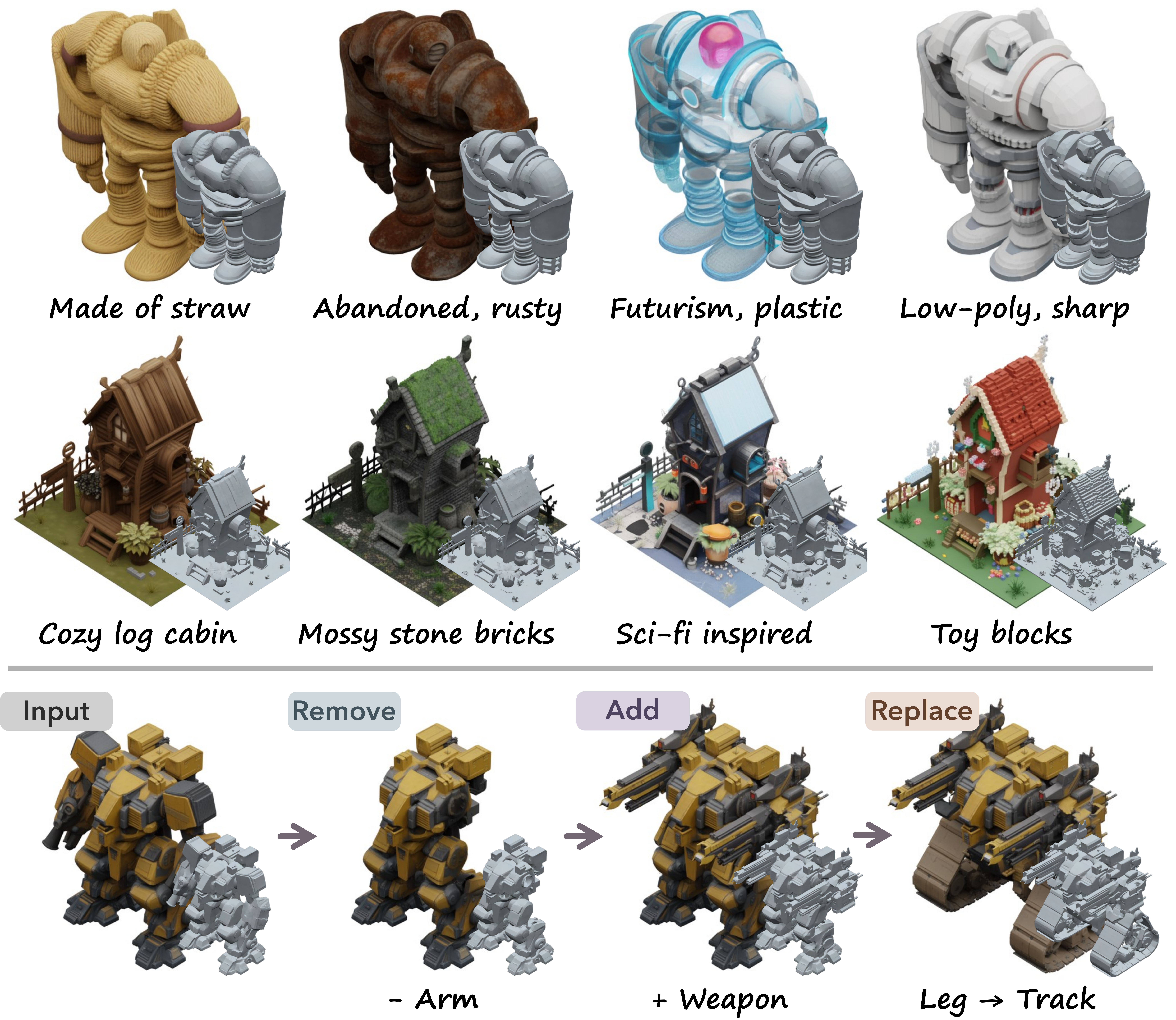}
    \end{minipage} 
    \vspace{-8pt}
	\caption{\textbf{Top:} Given coarse structures, our method generates 3D asset variations coherent with the text prompts. \textbf{Bottom:} Tuning-free region-specific editing results of our method, guided by text or image prompts. More results in Fig.\ref{fig:teaser} and Sec.~\ref{sec:more_results}.}
	\label{fig:applications}
    \vspace{-8pt}
\end{figure}

\paravspace
\paragraph{3D asset variations.}
Figure~\ref{fig:teaser} and~\ref{fig:asset_variation} show 3D asset variation results. Our method produces variants adhering to the overall shape of the given structures while exhibiting diverse appearance and geometry details guided by the text.

\paravspace
\paragraph{Region-specific editing of 3D assets.} Figure~\ref{fig:teaser} and~\ref{fig:local_editing} illustrate the editing sequences of two 3D assets, involving removal, addition, and replacement operations. Corresponding prompts (either text or image) for each step are provided. Our method enables detailed local region editing, such as adding a river and bridge in the island example.

\section{Conclusion}\label{sec:conclusion}
We introduced a novel 3D generation method for versatile
and high-quality 3D asset creation. At its core lies \textsc{SLat}, a structured latent representation that allows decoding to versatile output formats by comprehensively encoding both geometry and appearance information into localized latents anchored on a sparse 3D grid, where the latents are fused and processed from dense multiview image features extracted by a powerful vision foundation model. We proposed a two-stage generation pipeline utilizing rectified flow transformers tailored for \textsc{SLat} generation at scale. Extensive experiments demonstrated the superiority of our method in 3D generation, in terms of quality, versatility, and editability, highlighting its strong potential for a wide range of real-world applications in digital production.


{\small
\bibliographystyle{ieeenat_fullname}
\bibliography{ref}
}

\clearpage
\appendix
\twocolumn[\begin{center}
   \Large \textbf{Structured 3D Latents for Scalable and Versatile 3D Generation\\ \emph{(Supplementary Material)}}
\end{center}]

\begin{strip}
    \centering  
    \scriptsize
    \captionsetup{type=table,font=small,position=top}
    \caption{Network configurations used in this paper. \emph{SW} stands for ``Shifted Window", \emph{MSA} and \emph{MCA} for ``Multihead Self-Attention" and ``Multihead Cross-Attention", and \emph{Sp. Conv.} for ``Sparse Convolution".}  
    \vspace{-8pt}
	\begin{tabular}{c|ccccccc}  
        \toprule
        \textbf{Network} & \textbf{\#Layer} & \textbf{\#Dim.} & \textbf{\#Head} & \textbf{Block Arch.} & \textbf{Special Modules} & \textbf{\#Param.} \\
        \midrule  
        $\boldsymbol{\mathcal{E}}_\mathrm{S}$ & -- & -- & -- & -- & 3D Conv. U-Net & 59.3M \\
        $\boldsymbol{\mathcal{D}}_\mathrm{S}$ & -- & -- & -- & -- & 3D Conv. U-Net & 73.7M \\
        $\boldsymbol{\mathcal{E}}$ & 12 & 768 & 12 & 3D-SW-MSA + FFN & 3D Swin Attn. & 85.8M \\
        $\boldsymbol{\mathcal{D}}_\mathrm{GS}$ & 12 & 768 & 12 & 3D-SW-MSA + FFN & 3D Swin Attn. & 85.4M \\
        $\boldsymbol{\mathcal{D}}_\mathrm{RF}$ & 12 & 768 & 12 & 3D-SW-MSA + FFN & 3D Swin Attn. & 85.4M \\
        $\boldsymbol{\mathcal{D}}_\mathrm{M}$ & 12 & 768 & 12 & 3D-SW-MSA + FFN & 3D Swin Attn. + Sp. Conv. Upsampler & 90.9M \\
        $\boldsymbol{\mathcal{G}}_\mathrm{S}$-B (text ver.) & 12 & 768 & 12 & MSA + MCA + FFN & QK Norm. & 157M \\
        $\boldsymbol{\mathcal{G}}_\mathrm{S}$-L (text ver.) & 24 & 1024 & 16 & MSA + MCA + FFN & QK Norm. & 543M \\
        $\boldsymbol{\mathcal{G}}_\mathrm{S}$-XL (text ver.) & 28 & 1280 & 16 & MSA + MCA + FFN & QK Norm. & 975M \\
        $\boldsymbol{\mathcal{G}}_\mathrm{S}$-L (image ver.) & 24 & 1024 & 16 & MSA + MCA + FFN & QK Norm. & 556M \\
        $\boldsymbol{\mathcal{G}}_\mathrm{L}$-B (text ver.) & 12 & 768 & 12 & MSA + MCA + FFN & QK Norm. + Sp. Conv. Downsampler / Upsampler + Skip Conn. & 185M \\
        $\boldsymbol{\mathcal{G}}_\mathrm{L}$-L (text ver.) & 24 & 1024 & 16 & MSA + MCA + FFN & QK Norm. + Sp. Conv. Downsampler / Upsampler + Skip Conn. & 588M \\
        $\boldsymbol{\mathcal{G}}_\mathrm{L}$-XL (text ver.) & 28 & 1280 & 16 & MSA + MCA + FFN & QK Norm. + Sp. Conv. Downsampler / Upsampler + Skip Conn. & 1073M \\
        $\boldsymbol{\mathcal{G}}_\mathrm{L}$-L (image ver.) & 24 & 1024 & 16 & MSA + MCA + FFN & QK Norm. + Sp. Conv. Downsampler / Upsampler + Skip Conn. & 600M \\
        \bottomrule
    \end{tabular}  
    \label{tab:network} 
\end{strip}

\section{More Implementation Details} \label{sec:implement}
\subsection{Network Architectures}
The networks used in our method primarily consist of transformers~\cite{vaswani2017attention}, augmented by a few specialized modules. The configurations and statistics for each network are listed in Tab.~\ref{tab:network}. In particular, $\boldsymbol{\mathcal{E}}_\mathrm{S}$ and $\boldsymbol{\mathcal{D}}_\mathrm{S}$ compose the VAE designed for sparse structures, as discussed in Sec.~\ref{sec:generation} in the main paper. The remaining networks are also defined in the main paper. Below, we provide detailed descriptions of the architectures of the specialized modules introduced.

\paravspace
\paragraph{3D convolutional U-net.}
The VAE for sparse structures ($\boldsymbol{\mathcal{E}}_\mathrm{S}$ and $\boldsymbol{\mathcal{D}}_\mathrm{S}$) is introduced to enhance the efficiency of the structure generator $\boldsymbol{\mathcal{G}}_\mathrm{S}$ and to convert the binary grids of active voxels into continuous latents for flow training. Its architecture is similar to the VAEs in LDM~\cite{rombach2022high}, but it employs 3D convolutions and omits self-attention metchanisms. $\boldsymbol{\mathcal{E}}_\mathrm{S}$ ($\boldsymbol{\mathcal{D}}_\mathrm{S}$) consists of a series of residual blocks and downsampling (upsampling) blocks, reducing the spatial size from $64^3$ to $16^3$. The feature channels are set to $32$, $128$, $512$ for spatial sizes of $64^3$, $32^3$, $16^3$, respectively. The latent channel dimension is set to $8$. We utilize pixel shuffle~\cite{shi2016real} in the upsampling block and replace group normalizations with layer normalizations.

\paravspace
\paragraph{3D shifted window attention.}
In the VAE for structured latents (\textsc{SLat}), we employ 3D shifted window attention to facilitate local information interaction and improve efficiency. Specifically, we partition the $64^3$ space into $8^3$ windows, with tokens inside each window performing self-attention independently. Despite the potential variation in the number of tokens per window, this challenge can be efficiently addressed using modern attention implementations (\eg, FlashAttention~\cite{dao2023flashattention} and xformers~\cite{xFormers2022}). The transformer blocks alternate between non-shifted window attention and window attention shifted by $(4,4,4)$, ensuring that the windows in adjacent layers overlap uniformly.

\paravspace
\paragraph{QK normalization.}
Similar to the challenges reported in SD3~\cite{esser2024scaling}, we encounter training instability caused by the exploding norms of queries and keys within the multi-head attention blocks. To mitigate this issue, we follow~\cite{esser2024scaling} to apply root mean square normalizations~\cite{zhang2019root} (RMSNorm) to the queries and keys before sending them into the attention operators.

\paravspace
\paragraph{Sparse convolutional downsampler/upsampler.}
In $\boldsymbol{\mathcal{D}}_\mathrm{M}$ and $\boldsymbol{\mathcal{G}}_\mathrm{L}$, it is necessary to alter the spatial size of sparse tensors to increase the resolution of the SDF grid for meshes and to improve the efficiency of the \textsc{SLat} generator, respectively. To achieve this, we employ downsampling and upsampling blocks equipped with sparse convolutions~\cite{wang2017ocnn}. These blocks are composed of residual networks with two sparse convolutional layers, skip connections with optional linear mappings, and pooling or unpooling operators. We use average pooling and nearest-neighbor unpooling. For $\boldsymbol{\mathcal{G}}_\mathrm{L}$, given that the structures of $64^3$ are pre-determined, we only average the features from active voxels within each $2^3$ pooling window and recover the $64^3$ structures during unpooling. This is done by assigning values to active voxels from their nearest neighbors in the $32^3$ space. For $\boldsymbol{\mathcal{D}}_\mathrm{M}$, we simply subdivide each voxel into $2^3$, resulting in a new sparse tensor with doubled spatial dimensions in each upsampling block.

\subsection{Training Details}\label{sec:training_details}

We provide more details about the training process for each model, including hyperparameter tuning, algorithm details, and loss function designs.

\paravspace
\paragraph{Sparse structure VAE.}  
We frame the training of the sparse structure VAE as a binary classification problem, given the binary nature of the active voxels. Each decoded voxel is classified as either positive (active) or negative (inactive). Due to the imbalance between positive and negative labels, where active voxels are sparser than inactive ones, we adopt the Dice loss~\cite{milletari2016v} to effectively manage this disparity. 

\paravspace
\paragraph{Structured latent VAE.}
For the versatile decoding of \textsc{SLat}, we implement decoders for various 3D representations, namely $\boldsymbol{\mathcal{D}}_\mathrm{GS}$ for 3D Gaussians~\cite{kerbl20233d}, $\boldsymbol{\mathcal{D}}_\mathrm{RF}$ for Radiance Fields~\cite{mildenhall2021nerf}, and $\boldsymbol{\mathcal{D}}_\mathrm{M}$ for meshes. We provide detailed information on their respective training processes.

\vspace{4pt}
\noindent{\textit{(a) 3D Gaussians.}}
Following Mip-Splatting~\cite{yu2024mip}, we address aliasing by setting the minimal scale for Gaussians to $9e-4$ and the variance of the screen space Gaussian filter to $0.1$. The value $9e-4$ is derived from the assumption of  a $512^3$ sampling rate within the $(-0.5,0.5)^3$ cube. For each active voxel, 32 Gaussians are predicted (\ie, $K=32$ in the main paper). Since original density control schemes are not applicable when Gaussians are predicted by neural networks, we employ regularizations for volume~\cite{lombardi2021mixture} and opacity of the Gaussians to prevent their degeneration, specifically to avoid them becoming excessively large or transparent. The full training objective is:
\begin{equation}
        \mathcal{L}_{\mathrm{GS}}=\mathcal{L}_{\mathrm{recon}}+\mathcal{L}_{\mathrm{vol}}+\mathcal{L}_{\alpha},
\end{equation}
where $\mathcal{L}_{\mathrm{recon}}$, $\mathcal{L}_{\mathrm{vol}}$ and $\mathcal{L}_{\alpha}$ are defined below:
\begin{equation}
    \begin{split}
    \mathcal{L}_\mathrm{{recon}}=\mathcal{L}_1&+0.2(1-\mathrm{SSIM})+0.2{\mathrm{LPIPS}},\\
\mathcal{L}_{\mathrm{vol}}&=\frac1{LK}\sum_{i=1}^{L}\sum_{k=1}^K\prod\boldsymbol{s}_i^k,\\      
\mathcal{L}_{\alpha}&=\frac1{LK}\sum_{i=1}^{L}\sum_{k=1}^K(1-\alpha_i^k)^2.
    \end{split} \label{eq:recon}
\end{equation}

\vspace{4pt}
\noindent{\textit{(b) Radiance Fields.}}
We predict 4 orthogonal vectors $\boldsymbol{v}_i^\mathrm{x},\boldsymbol{v}_i^\mathrm{y},\boldsymbol{v}_i^\mathrm{z},\boldsymbol{v}_i^\mathrm{c}$ for each active voxel. These vectors represent the CP-decomposition~\cite{chen2022tensorf} of a local $8^3$ radiance volume $\boldsymbol{V}\in\mathbb{R}^{8\times8\times8\times4}$:
\begin{equation}
    \boldsymbol{V}_{i,xyzc}=\sum_{r=1}^{R}\boldsymbol{v}_{i,rx}^\mathrm{x}\boldsymbol{v}_{i,ry}^\mathrm{y}\boldsymbol{v}_{i,rz}^\mathrm{z}\boldsymbol{v}_{i,rc}^\mathrm{c}.
\end{equation}
The last dimension of $\boldsymbol{V}$, which has a size of 4, contains the color and density information. We set the rank $R=16$. The recovered local volumes are then assembled according to the position of their respective active voxels, forming a $512^3$ radiance field. Additionally, we implement an efficient differentiable renderer using CUDA, which enables real-time rendering by integrating sorting, ray marching, radiance integration, and the CP reconstruction into a single kernel. The training objective of $\boldsymbol{\mathcal{D}}_\mathrm{RF}$ is $\mathcal{L}_{\mathrm{recon}}$ as defined in Eq.~\eqref{eq:recon}.

\vspace{4pt}
\noindent{\textit{(c) Meshes.}} We increase the spatial size of sparse structures from $64^3$ to $256^3$, by appending two aforementioned sparse convolutional upsamplers after the transformer backbone. For $\boldsymbol{\mathcal{D}}_\mathrm{M}$, although our primary focus is on shape (geometry), we also predict colors and normal maps for the meshes. As a result, the final output for each high-resolution active voxel is:
\begin{equation}
    (\boldsymbol{w}_i^j, \boldsymbol{d}_i^j, \boldsymbol{c}_i^j, \boldsymbol{n}_i^j).
\end{equation}
Here, $\boldsymbol{w}_i^j = (\boldsymbol{\alpha}_i^j, \boldsymbol{\beta}_i^j, \gamma_i^j, \boldsymbol{\delta}_i^j)$ are the flexible parameters defined in FlexiCubes~\cite{shen2023flexicubes}, where $\boldsymbol{\alpha}_i^j \in\mathbb{R}^{8}$ and $\boldsymbol{\beta}_i^j \in\mathbb{R}^{12}$ are interpolation weights per voxel, $\gamma_i^j \in\mathbb{R}$ is the splitting weights per voxel, and $\boldsymbol{\delta}_i^j \in\mathbb{R}^{8\times3}$ is per vertex deformation vectors of the voxel. In addition, $\boldsymbol{d}_i^j \in\mathbb{R}^{8}$ is the signed distance values for the eight vertices of the voxel, $\boldsymbol{c}_i^j \in\mathbb{R}^{8\times3}$ denotes vertex colors, and $\boldsymbol{n}_i^j \in\mathbb{R}^{8\times3}$ represents vertex normals. Since each vertex is connected to multiple voxels, we derive the final vertex attributes (\ie, $\boldsymbol{\delta}$, $\boldsymbol{d}$, $\boldsymbol{c}$, and $\boldsymbol{n}$) by averaging the predictions from all associated voxels.
    
To simplify implementation, we attach the sparse structure to a dense grid for differentiable surface extraction using FlexiCubes. For all inactive voxels in the dense grid, we set their signed distance values to $1.0$ and all other associated attributes to zero.
We then extract meshes from the 0-level iso-surfaces of the dense grid. For each mesh vertex, its associated attributes (\ie, $\boldsymbol{c}$ and $\boldsymbol{n}$) are interpolated from those of the corresponding grid vertices. 
We utilize Nvdiffrast~\cite{laine2020modular} to render the extracted mesh along with its attributes, producing a foreground mask $\boldsymbol{M}$, a depth map $\boldsymbol{D}$, a normal map $\boldsymbol{N}_m$ directly derived from the mesh, an RGB image $\boldsymbol{C}$, and a normal map $\boldsymbol{N}$ from the predicted normals. The training objective is then defined as follows:
\begin{equation}
\mathcal{L}_{\mathrm{M}}=\mathcal{L}_\mathrm{geo}+0.1\mathcal{L}_\mathrm{color}+\mathcal{L}_{\mathrm{reg}},
\end{equation}
where $\mathcal{L}_\mathrm{geo}$ and $\mathcal{L}_\mathrm{color}$ are written as:
\begin{equation}
\begin{split}
\mathcal{L}_\mathrm{geo} = \mathcal{L}_1(\boldsymbol{M})+&10 \mathcal{L}_{\mathrm{Huber}}(\boldsymbol{D})+\mathcal{L}_{\mathrm{recon}}(\boldsymbol{N}_{m}),\\
\mathcal{L}_\mathrm{color} = &\mathcal{L}_{\mathrm{recon}}(\boldsymbol{C}) + \mathcal{L}_{\mathrm{recon}}(\boldsymbol{N}).
\end{split}
\end{equation}
Here, $\mathcal{L}_{\mathrm{recon}}$ is defined identically to Eq.~\eqref{eq:recon}. Finally, $\mathcal{L}_{\mathrm{reg}}$ consists of three terms:
\begin{equation}
    \mathcal{L}_\mathrm{reg} = \mathcal{L}_{\mathrm{consist}} +\mathcal{L}_{\mathrm{dev}} + 0.01\mathcal{L}_{\mathrm{tsdf}},
\end{equation}
where $\mathcal{L}_{\mathrm{consist}}$ penalizes the variance of attributes associated with the same voxel vertex, $\mathcal{L}_{\mathrm{dev}}$ is a regularization term defined in FlexiCubes to ensure plausible mesh extraction, and $\mathcal{L}_{\mathrm{tsdf}}$ enforces the predicted signed distance values $\boldsymbol{d}$ to closely match the distances between grid vertices and the extracted mesh surface, helping to stablize the training process in its early stages.

\paravspace
\paragraph{Rectified flow models.}
We employ rectified flow models $\boldsymbol{\mathcal{G}}_\mathrm{S}$ and $\boldsymbol{\mathcal{G}}_\mathrm{L}$ for sparse structure generation and structured latent generation, respectively. During training, we alter the timestep sampling distribution, replacing the $\mathrm{logitNorm}(0,1)$ distribution used in SD3 with $\mathrm{logitNorm}(1,1)$. We evaluate their performance at each stage of our generation pipeline using the Toys4k dataset. As shown in Tab.~\ref{tab:ablation_timestep}, the latter provides a better fit for our task and we set it as the default setting.

\begin{figure*}[t]
	\small
	\setlength\tabcolsep{1pt}
	\centering
	\begin{tabular}{ccccc}
		\includegraphics[width=0.2\linewidth]{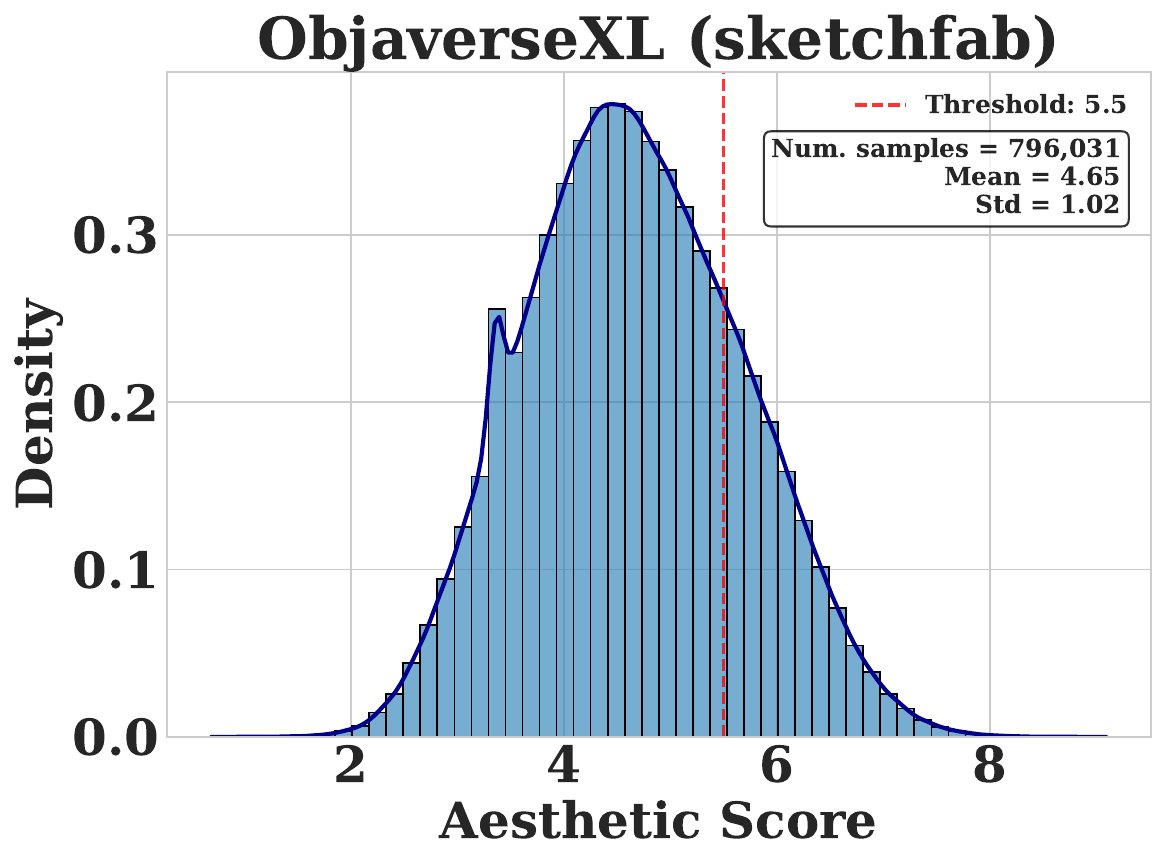} & \includegraphics[width=0.2\linewidth]{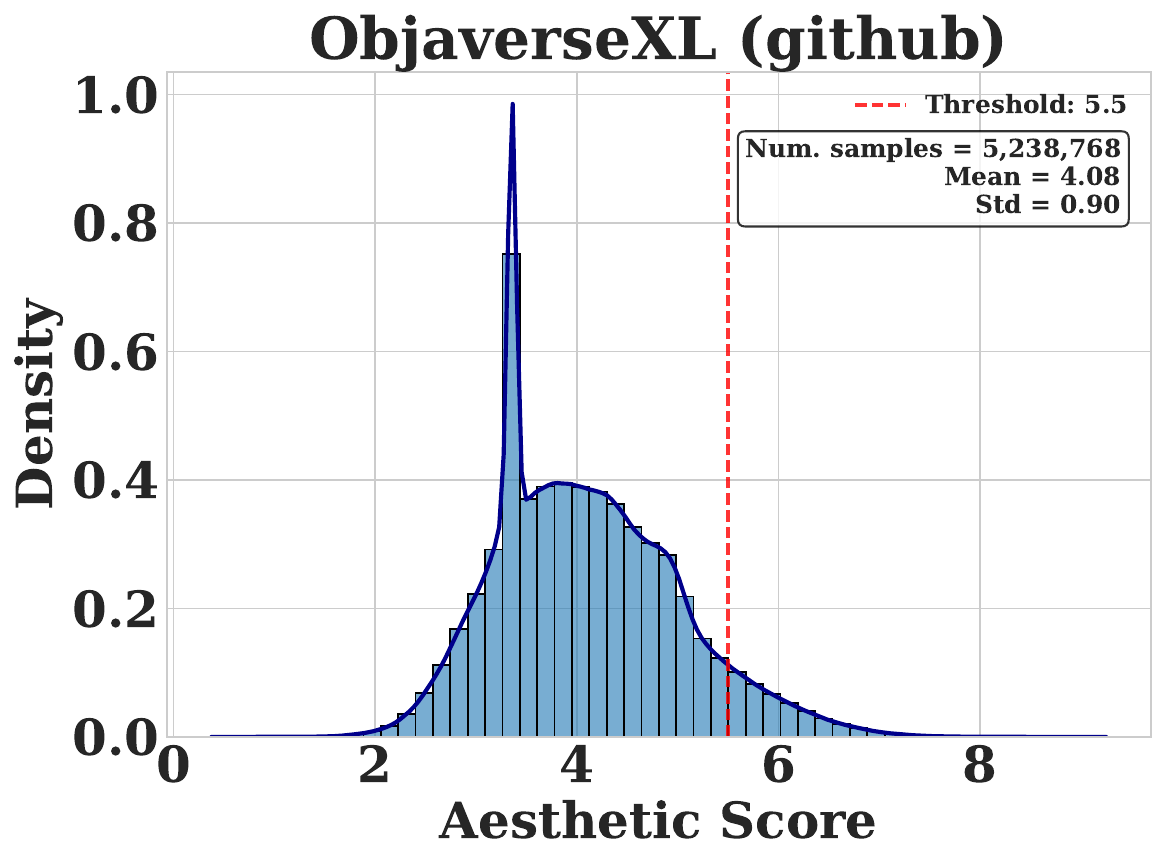} & \includegraphics[width=0.2\linewidth]{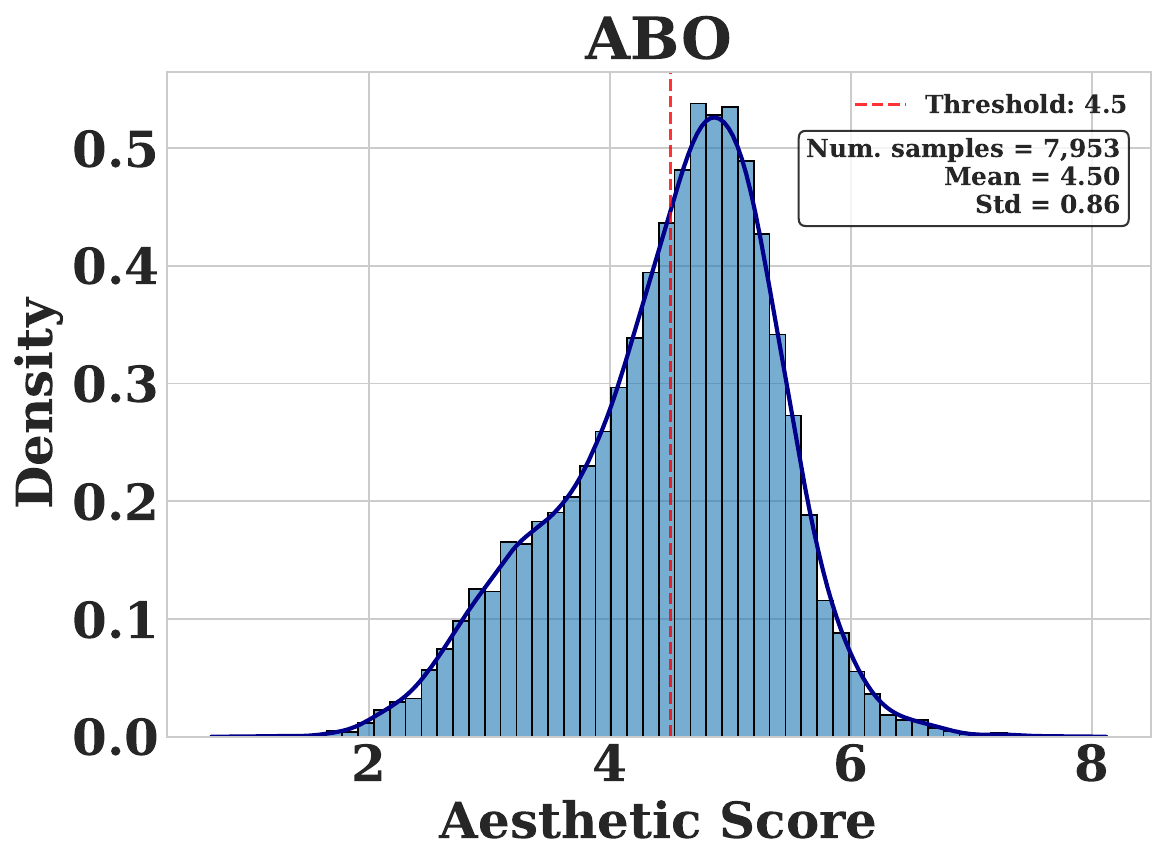} & \includegraphics[width=0.2\linewidth]{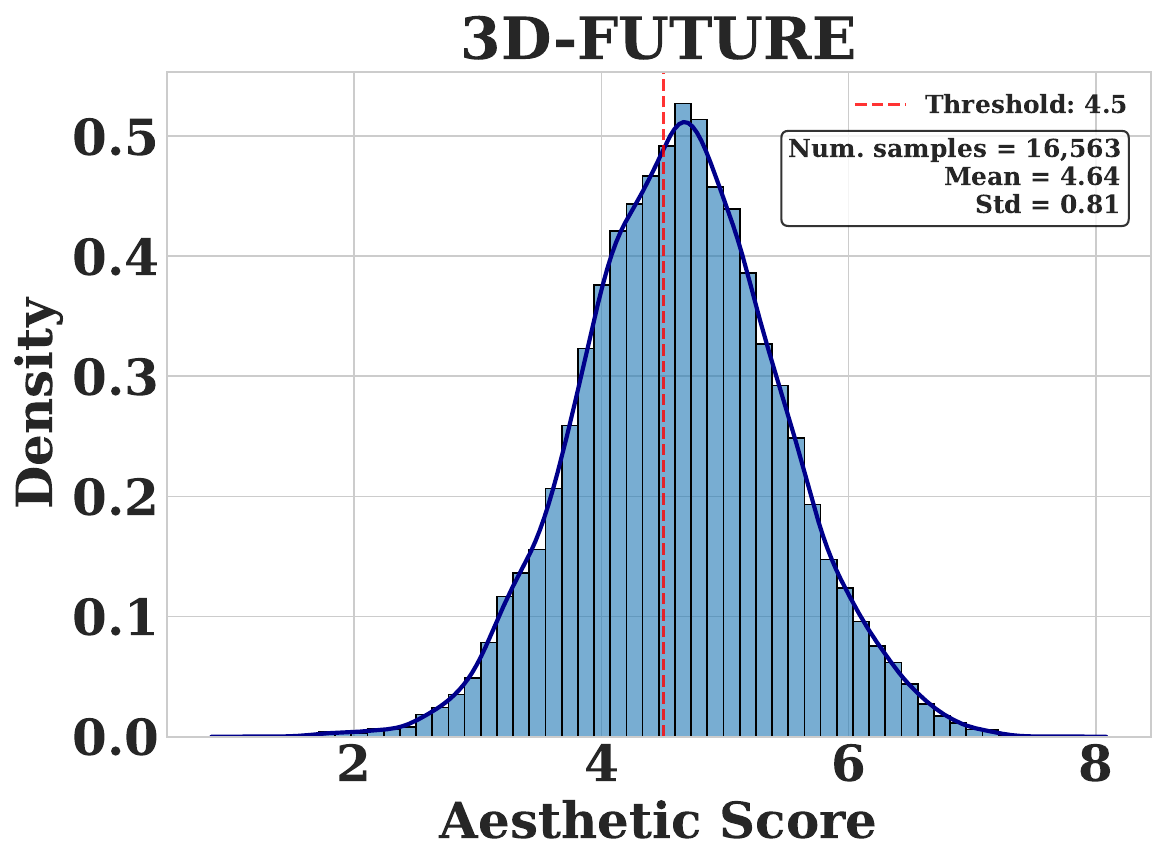} &
        \includegraphics[width=0.2\linewidth]{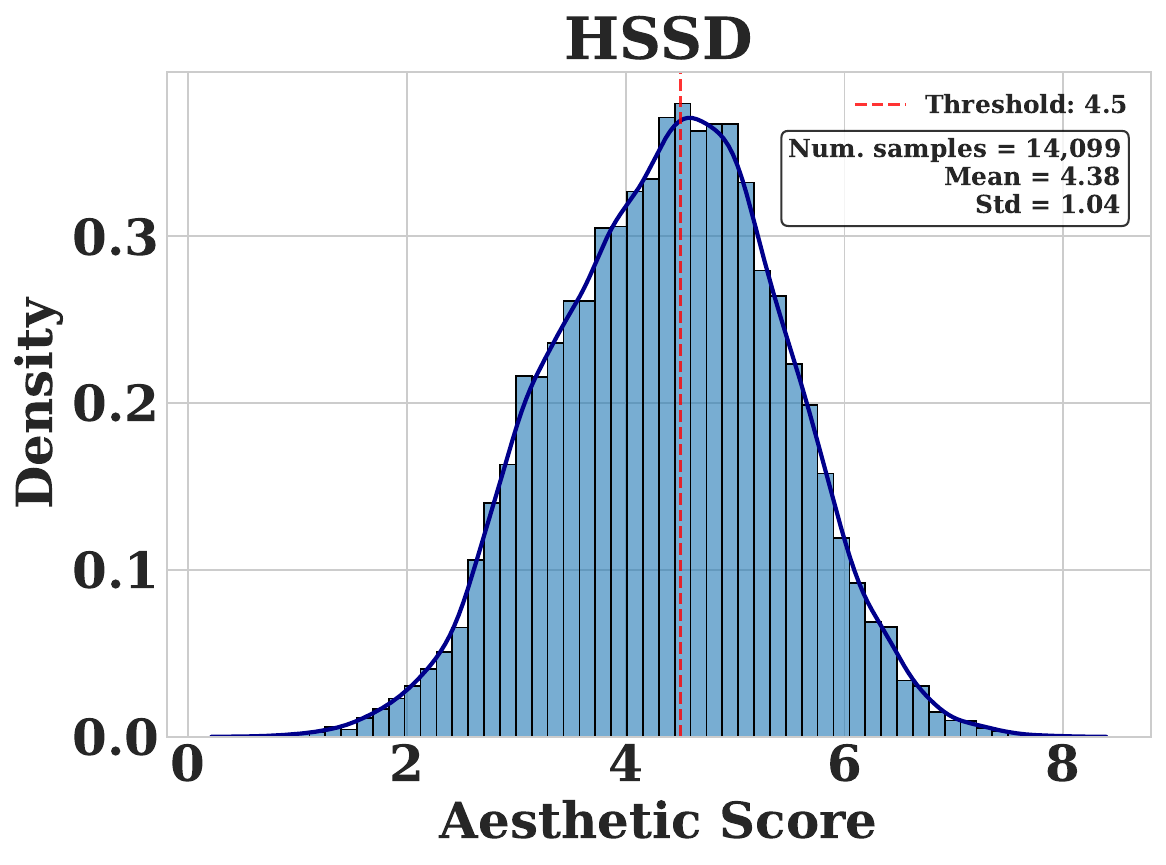} 
	\end{tabular}
	\caption{Distribution of aesthetic scores in each dataset.}
	\label{fig:aesthetic_scores_distribution}
\end{figure*}

\begin{table}[t]  
	\centering  
	\scriptsize
    \caption{Ablation study on timestep sampling distributions.}
    \vspace{-8pt}
	\setlength{\tabcolsep}{4pt}
    \begin{tabular}{cc|cc}  
		\toprule
		& \textbf{Distribution} & $\textbf{CLIP}\!\uparrow$ & $\textbf{FD}_\textbf{dinov2}\!\downarrow$ \\  
		\midrule
        \multirow{2}{*}{Stage 1}
        & $\mathrm{logitNorm}(0,1)$ &  26.03 &  287.33  \\
        & $\mathrm{logitNorm}(1,1)$ & \textbf{26.37} &  \textbf{269.56} \\
		\midrule
        \multirow{2}{*}{Stage 2}
        & $\mathrm{logitNorm}(0,1)$ & \textbf{26.61} & 242.36 \\
        & $\mathrm{logitNorm}(1,1)$ & \textbf{26.61} & \textbf{240.20} \\
		\bottomrule  
	\end{tabular} 
    \vspace{-8pt}
	\label{tab:ablation_timestep}  
\end{table} 

\section{Data Preparation Details}

Recognizing the critical importance of both the quantity and quality of training data for scaling up the generative models, we carefully curate our training data from currently available open-source 3D datasets to construct a high-quality, large-scale 3D dataset. Moreover, we employed state-of-the-art multimodal model, GPT4o~\cite{2024GPT4o}, to caption each 3D asset, ensuring precise and detailed text descriptions. This facilitates accurate and controllable generation of 3D assets from text prompts. In the following sections, we will first briefly introduce each 3D dataset utilized, and then provide details about our data curation pipeline. In addition, we provide a comprehensive explanation of both the captioning process and our rendering settings.

\subsection{3D Datasets}

\paragraph{Objaverse-XL~\cite{deitke2024objaverse}.} Objaverse-XL is the largest open-source 3D dataset, comprising over 10 million 3D objects sourced from diverse platforms such as GitHub, Thingiverse, Sketchfab, Polycam, and the Smithsonian Institution. This extensive collection includes manually designed objects, photogrammetry scans of landmarks and everyday items, as well as professional scans of historic and antique artifacts. Despite its large scale, Objaverse-XL is quite noisy, containing a significant number of low-quality objects, such as those with missing parts, low-resolution textures, and simplified geometries. Therefore, we include only the objects from Sketchfab (also known as ObjaverseV1~\cite{deitke2023objaverse}) and GitHub in our training dataset and perform a thorough filtering process to clean the dataset.

\paravspace
\paragraph{ABO~\cite{collins2022abo}.} ABO includes about 8K high-quality 3D models provided by Amazon.com. These models are designed by artists and feature complex geometries and high-resolution materials. The dataset encompasses 63 categories, primarily focusing on furniture and interior decoration.

\paravspace
\paragraph{3D-FUTURE~\cite{fu20213d}.} 3D-FUTURE contains around 16.5K 3D models created by experienced designers for industrial production, offering rich geometric details and informative textures. This dataset specifically focuses on 3D furniture shapes designed for household scenarios.

\paravspace
\paragraph{HSSD~\cite{khanna2023hssd}.}
HSSD is a high-quality, human-authored synthetic 3D scene dataset designed to test navigation agent generalization to realistic 3D environments. It includes a total of 14K 3D models, primarily assets of indoor scenes such as furniture and decorations.

\paravspace
\paragraph{Toys4k~\cite{stojanov2021using}.} Toys4k contains approximately 4K high-quality 3D objects from 105 object categories, featuring a diverse set of object instances within each category. Since previous works have not utilized this dataset for training, we leverage it as our testing dataset to evaluate the generalization of our model.

\begin{figure}[t]
	\small
	\centering
	\begin{tabular}{cccc}
		\includegraphics[width=0.2\linewidth]{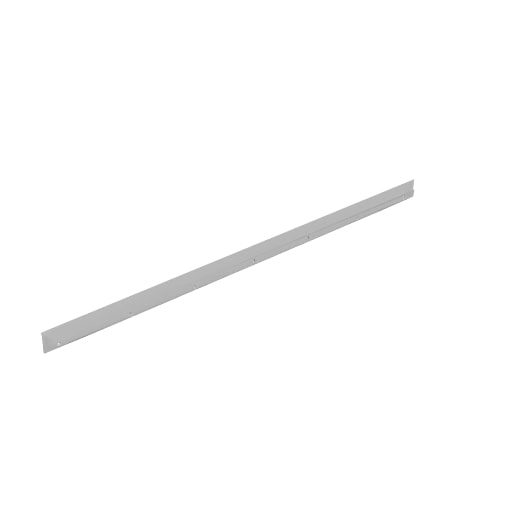} & \includegraphics[width=0.2\linewidth]{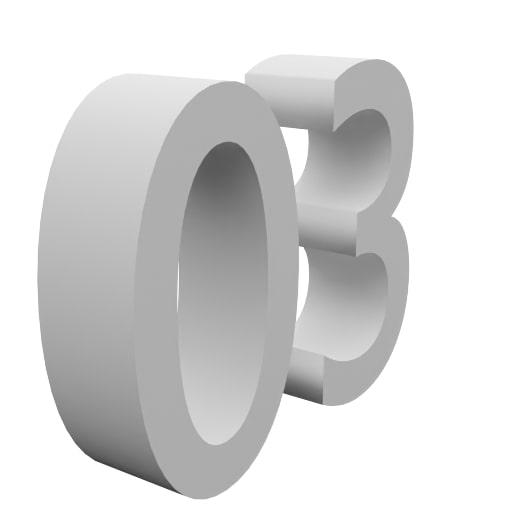} & \includegraphics[width=0.2\linewidth]{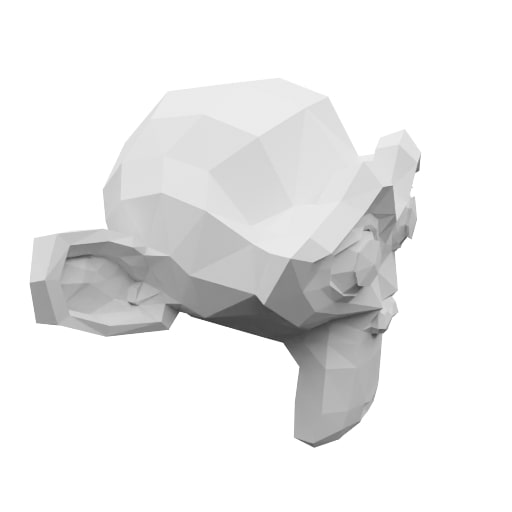} & \includegraphics[width=0.2\linewidth]{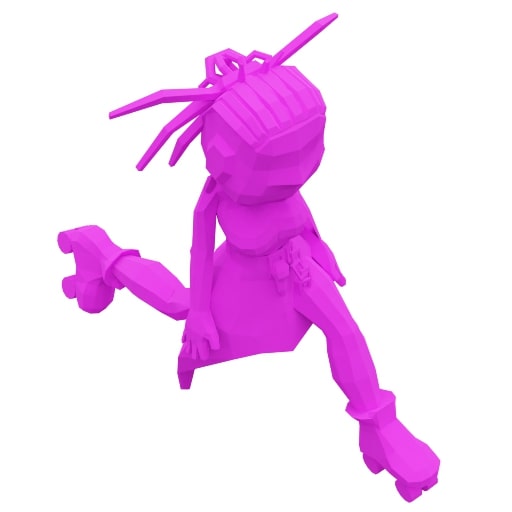} \\
        Score: 2.32 & Score: 3.84 & Score: 4.91 & Score: 5.24 \\
        \includegraphics[width=0.2\linewidth]{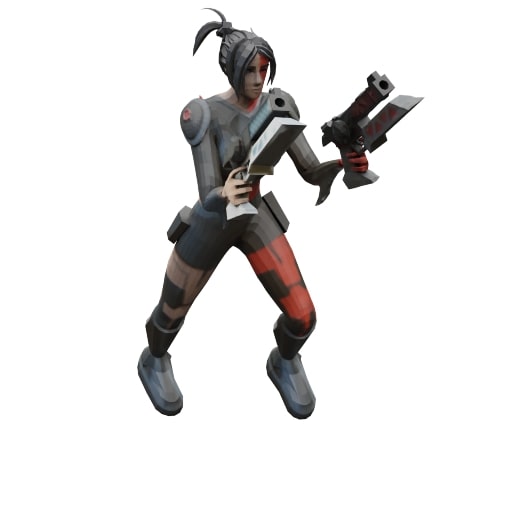} & \includegraphics[width=0.2\linewidth]{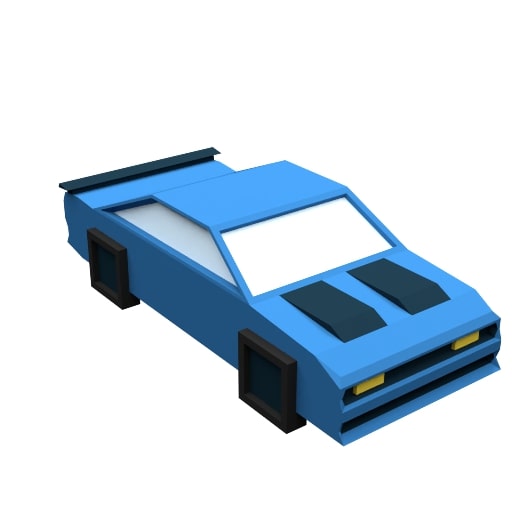} & \includegraphics[width=0.2\linewidth]{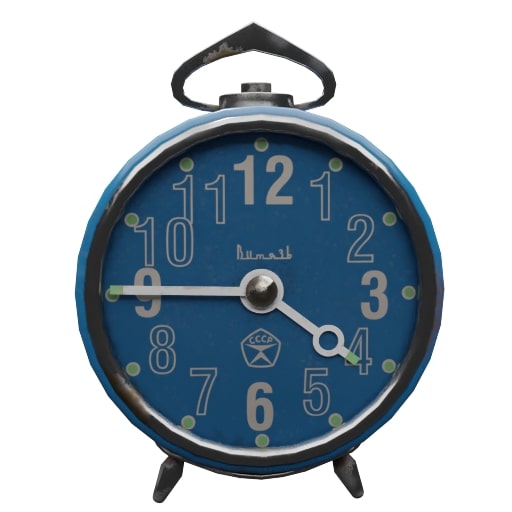} & \includegraphics[width=0.2\linewidth]{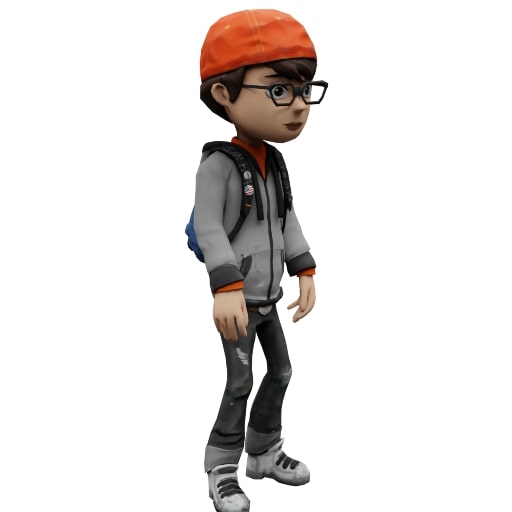} \\
        Score: 5.85 & Score: 6.04 & Score: 6.29 & Score: 7.03 \\
	\end{tabular}
	\caption{3D asset examples from Objaverse-XL with their corresponding aesthetic scores.}
	\label{fig:aesthetic_scores}
\end{figure}

\subsection{Data Curation Pipeline}

To ensure high-quality training data, we implement a systematic curation process. First, we render 4 images from uniformly distributed viewpoints around each 3D object. We then employ a pretrained aesthetic assessment model~\footnote{\href{https://github.com/christophschuhmann/improved-aesthetic-predictor}{https://github.com/christophschuhmann/improved-aesthetic-predictor}} to evaluate the quality of each 3D asset. More specifically, we assess the average aesthetic score across 4 rendered view for each 3D object. We empirically find this scoring mechanism can effectively identify objects with poor visual quality -- those that receive low aesthetic scores typically exhibit undesirable characteristics such as minimal texturing or overly simplistic geometry. We visualize the distribution of aesthetic scores in each dataset in Fig.~\ref{fig:aesthetic_scores_distribution}, and further provide some examples in Fig.~\ref{fig:aesthetic_scores} to illustrate the correspondance between the quality of 3D assets and their aesthetic scores. By filtering out objects with average aesthetic score below a certain aesthetic score threshold (\ie, 5.5 for Objaverse-XL and 4.5 for the other datasets), we maintain a high standard of geometric and textural complexity in our dataset. After filtering, there are about 500K high-quality 3D objects left (more details listed in Tab.~\ref{tab:dataset}), which comprise our training dataset.

\begin{table}[t]
    \centering  
    \scriptsize
    \caption{Composition of the training set and evaluation set.}  
    \vspace{-8pt}
	\begin{tabular}{c|cc}  
        \toprule
        \textbf{Source} & \textbf{Aesthetic Score Threshold} & \textbf{Filtered Size} \\
        \midrule  
        ObjaverseXL (sketchfab) & 5.5 & 168307 \\
        ObjaverseXL (github) & 5.5 & 311843 \\
        ABO & 4.5 & 4485 \\
        3D-FUTURE & 4.5 & 9472 \\
        HSSD & 4.5 & 6670 \\
        \textbf{All (training set)} & -- & 500777 \\
        \midrule  
        Toys4k (evaluation set) & 4.5 & 3229 \\
        \bottomrule
    \end{tabular}  
    \label{tab:dataset} 
\end{table}

\subsection{Captioning Process}

\begin{figure}
    \centering
    \includegraphics[width=0.96\linewidth]{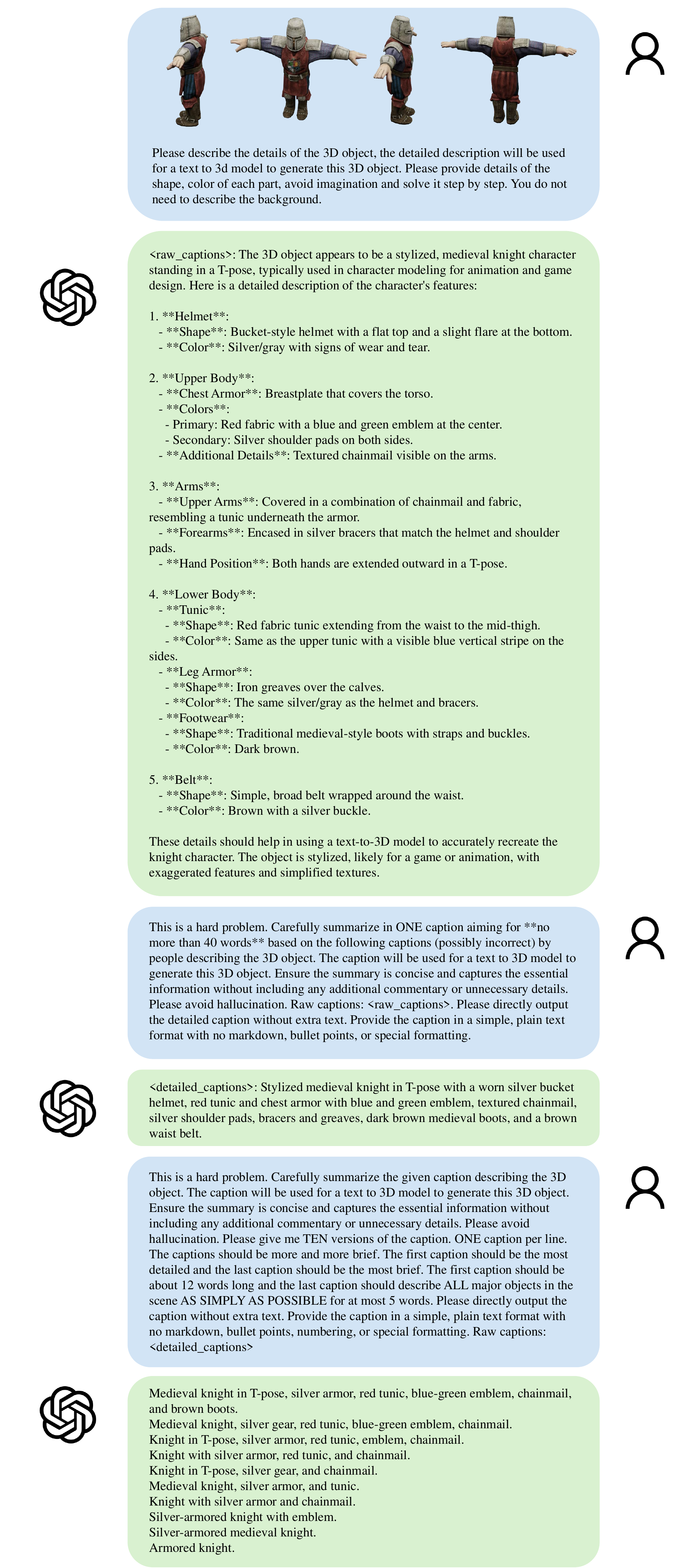}
    \vspace{-2pt}
    \caption{An example of our captioning process.}
    \label{fig:captioning_process}
\end{figure}

Current available captions~\cite{luo2024scalable} for 3D objects either suffer from poor alignment with the objects they describe or lack detailed descriptions~\cite{ge2024visual}, which hinders high-quality text-to-3D generation. Therefore, we carefully design a captioning process following~\cite{ge2024visual} to make the model generate precise and detailed text descriptions for each 3D object. To be more specific, we first employ GPT4o to produce a highly detailed description ``$<$raw\_captions$>$'' of the input rendered images. Subsequently, GPT4o distills the crucial information from ``$<$raw\_captions$>$'' into ``$<$detailed\_captions$>$'', typically comprising no more than 40 words. Additionally, we summarize the ``$<$detailed\_captions$>$'' into varying-length text prompts for augmentation in training. An illustration of the entire captioning process can be found in Fig.~\ref{fig:captioning_process}, which also includes the prompts designed for GPT4o.

\subsection{Rendering Process}

For VAE training, we sample 150 cameras looking at the origin with a FoV of $40^\circ$, uniformly distributed across a sphere with a radius of 2. We render the assets using Blender, with a smooth area lighting. For the image-conditioned generation model, we render a different set of images with augmented FoVs ranging from $10^\circ$ to $70^\circ$, which serves as image prompts during training.

\section{More Experiment Details}

\subsection{Evaluation Protocol}
In Sec.~\ref{sec:gen_results} and~\ref{sec:ablation} in the main paper, we conduct quantitative comparisons and ablation studies using a series of numerical metrics. We provide detailed protocols for their calculation below.

\paravspace
\paragraph{Reconstruction experiments.}
We randomly sample a subset of 500 instances from the filtered Toys4k dataset, which comprises 3,229 3D assets (see Tab.~\ref{tab:dataset}), as the evaluation set to assess the reconstruction fidelity of different latent representations. The evaluation is conducted in the following two aspects.

\vspace{4pt}
\noindent{\textit{(a) Appearance fidelity.}}
For each instance, we randomly sample one camera positioned on a sphere with a radius of 2, looking towards the origin with a FoV of $40^\circ$. We calculate PSNR and LPIPS between the rendered images from the reconstructed 3D assets and the ground truth images, and average the results as the final metrics. For 3DTopia-XL~\cite{chen20243dtopia}, which focuses on PBR materials, we report the reconstruction fidelity of albedo maps.

\vspace{4pt}
\noindent{\textit{(b) Geometry accuracy.}}
We employ Chamfer Distance (CD) and F-score of sampled point clouds to assess the overall geometry accuracy, as well as PSNR and LPIPS for rendered normal maps (\ie, PSNR-N and LPIPS-N) to evaluate surface details. Definitions for the point cloud metrics are listed below:
\begin{itemize}[leftmargin=2em]
    \item \emph{Chamfer Distance:}
    \begin{equation}
    \begin{split}
        \text{CD}(\boldsymbol{X},\boldsymbol{Y})&=\frac1{|\boldsymbol{X}|}\sum_{\boldsymbol{x}\in\boldsymbol{X}}\min_{\boldsymbol{y}\in\boldsymbol{Y}}\|\boldsymbol{x}-\boldsymbol{y}\|_2\\&+\frac1{|\boldsymbol{Y}|}\sum_{\boldsymbol{y}\in\boldsymbol{Y}}\min_{\boldsymbol{x}\in\boldsymbol{X}}\|\boldsymbol{y}-\boldsymbol{x}\|_2.
    \end{split}
    \end{equation}
    \item \emph{F-score:}
    \begin{equation}
    \begin{split}
        \text{FN} = \sum [\min_{\boldsymbol{y}\in \boldsymbol{Y}} &\| \boldsymbol{x} -\boldsymbol{y}\|_2 > r], \\
        \text{FP} = \sum [\min_{\boldsymbol{x}\in \boldsymbol{X}} &\|\boldsymbol{y} - \boldsymbol{x}\|_2 > r], \\
        \text{TP} = &|\boldsymbol{Y}| - \text{FP}, \\
        \text{precision} = &\frac{\text{TP}}{\text{TP}+\text{FP}},\\
        \text{recall} = &\frac{\text{TP}}{\text{TP}+\text{FN}},  \\
        \mathbf{F\text{-}score}(\boldsymbol{X}, \boldsymbol{Y})=&\frac{2\cdot\text{precision}\cdot \text{recall}}{\text{precision}+\text{recall}}.
    \end{split}
    \end{equation}
\end{itemize}
The point clouds used to assess the overall geometry accuracy (CD and F-score with $r=0.05$) are sampled from the outer surface of the reconstructed meshes. Specifically, we render depth maps for each mesh from 100 uniformly sampled views, with camera settings identical to that for appearance evaluation. The depth maps are then unprojected to 3D points. We randomly sample 100K points from all the 3D points as the point clouds for evaluation. 

For PSNR-N and LPIPS-N, as in the appearance metrics, we calculate the mean values across 500 image pairs (rendered results \emph{v.s.} ground truth), with one pair per instance. 

\paravspace
\paragraph{Generation experiments.}
For comparisons and ablation studies regarding generation quality, we utilize two evaluation sets: a subset of Toys4k with 1,250 randomly sampled instances and a subset of the training set with 5,000 instances. We employ Fréchet Distance (FD)~\cite{heusel2017gans} and Kernel Distance (KD)~\cite{binkowski2018demystifying} with various feature extractors (\ie, Inception-v3~\cite{szegedy2016rethinking}, DINOv2, and PointNet++~\cite{qi2017pointnet++}) to assess the overall quality of the generated outputs. Additionally, the CLIP score~\cite{radford2021learning} is used to evaluate the consistency between the generated results and the input prompts. For each prompt in the evaluation set, we generate one asset using the generation model and use these assets as the generated set for metrics calculation. We provide detailed calculations for each metric below.

\vspace{4pt}
\noindent{\textit{(a) Appearance quality.}} We employ 
    $\mathrm{FD}_\mathrm{incep}$, $\mathrm{KD}_\mathrm{incep}$, $\mathrm{FD}_\mathrm{dinov2}$, and $\mathrm{KD}_\mathrm{dinov2}$ as evaluation metrics. For each instance, we render 4 views using cameras with yaw angles of $\{0^\circ,90^\circ,180^\circ, 270^\circ\}$, and a pitch angle of $30^\circ$. All other camera settings are consistent with those in the reconstruction experiments. The rendered images are then used to calculate different metrics. For Toys4k, we use 5,000 images each for both the real and rendered sets, while for the training set, we use 20,000 images.

\vspace{4pt}
\noindent{\textit{(b) Geometry quality.}} 
    We utilize $\mathrm{FD}_\mathrm{point}$. Following Point-E~\cite{nichol2022point}, we prepare the point clouds by sampling 4,000 points from unprojected multiview depth maps using the farthest point sampling technique.

\vspace{4pt}
\noindent{\textit{(c) Prompt alignment.}} 
    We render 8 images per asset with yaw angles at every $45^\circ$, a pitch angle of $30^\circ$, and a radius of $2$. We calculate the cosine similarity between the CLIP features of images from the generated assets and their corresponding text or image prompts. The average of all similarities ($\times100$) is reported as the final CLIP score.

\begin{figure}
    \centering
    \includegraphics[width=\linewidth]{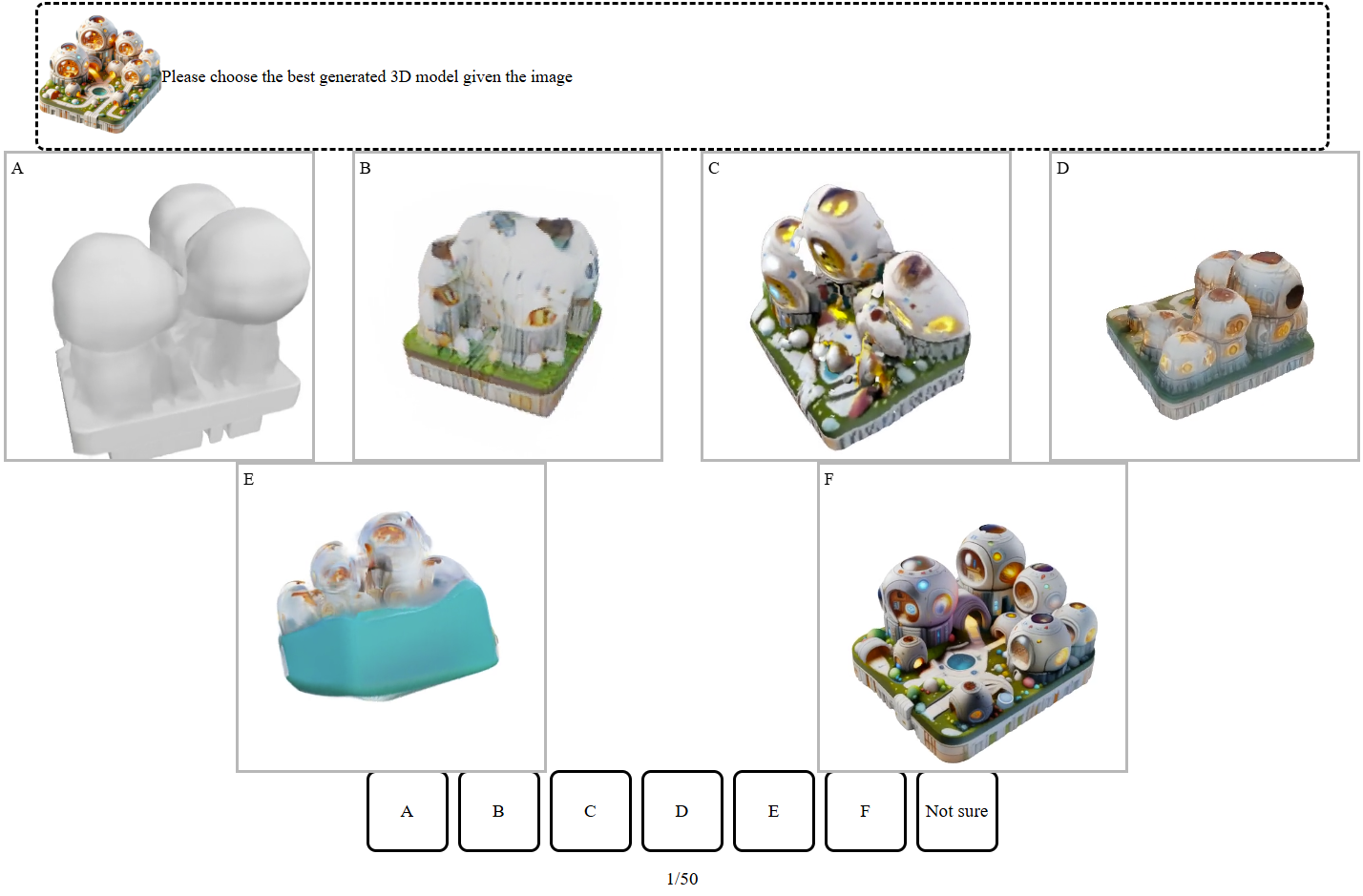}
    \vspace{-20pt}
    \caption{User interface used in our user study.}
    \label{fig:user_study_ui}
\end{figure}

\begin{table}[t]  
	\centering  
	\scriptsize
    \caption{Detailed statistics of the user study.}
    \vspace{-8pt}
	\setlength{\tabcolsep}{4pt}
	\begin{tabular}{c|cc|cc}  
		\toprule 
		\multirow{2}{*}{\textbf{Method}} & \multicolumn{2}{c|}{\textbf{Text-to-3D}} & \multicolumn{2}{c}{\textbf{Image-to-3D}} \\
        & $\textbf{Selections}\!\uparrow$ & $\textbf{Perentage}\!\uparrow$ & $\textbf{Selections}\!\uparrow$ & $\textbf{Perentage}\!\uparrow$ \\
		\midrule
        Not Sure & 56 & 4.2\% & 6 & 0.4\% \\
		Shap-E & 42 & 3.1\% & 6 & 0.4\% \\
		LGM & 70 & 5.2\% & 22 & 1.6\% \\
        InstantMesh & 123 & 9.1\% & 30 & 2.2\% \\
		3DTopia-XL & 5 & 0.4\% & 5 & 0.4\% \\
        Ln3Diff & 9 & 0.7\% & 6 & 0.4\% \\
		GaussianCube & 139 & 10.3\% & -- & -- \\
		\textbf{Ours} & \textbf{905} & \textbf{67.1\%} & \textbf{1277} & \textbf{94.5\%} \\
        \midrule
        \textbf{Total} & 1349 & 100\% & 1352 & 100\% \\
		\bottomrule
	\end{tabular}  
	\vspace{-8pt}
	\label{tab:user_study}  
\end{table} 

\subsection{User Study} \label{sec:user_study_detail}

We conducted a user study to evaluate the performance of various methods based on human preferences. Participants were presented with side-by-side comparisons of 3D assets generated by different methods. In each trial, they were given a text prompt or reference image, along with several rotating videos of candidate 3D assets generated using different techniques. The interface, as depicted in Fig.~\ref{fig:user_study_ui}, displayed the reference image at the top, followed by options representing the generated 3D models. Participants were asked to select the model that best matched the reference image in terms of visual fidelity and overall quality, or they could choose \textit{``Not sure"} if they were unable to make a decision. Each participant was assigned 50 trials, and their selections were recorded for analysis.

To ensure a diverse and unbiased evaluation, we implemented the following measures:
\begin{itemize}[leftmargin=2em]
\item The candidate 3D assets were not curated. Specifically, we sampled once per text or image prompt and used those samples directly in the study.
\item The 50 trials for each participant were randomly selected from a pool of 68 text-to-3D cases and 67 image-to-3D cases. The order of candidates in each trial was also randomized.
\end{itemize}

We collected responses from 104 participants. In total, 2,701 trials were answered, with an average of 25.97 responses each. Detailed statistics are in Tab.~\ref{tab:user_study}.

\section{More Results}\label{sec:more_results}

\subsection{3D Asset Generation}
\begin{figure*}[p]
    \centering
    \includegraphics[width=0.83\linewidth]{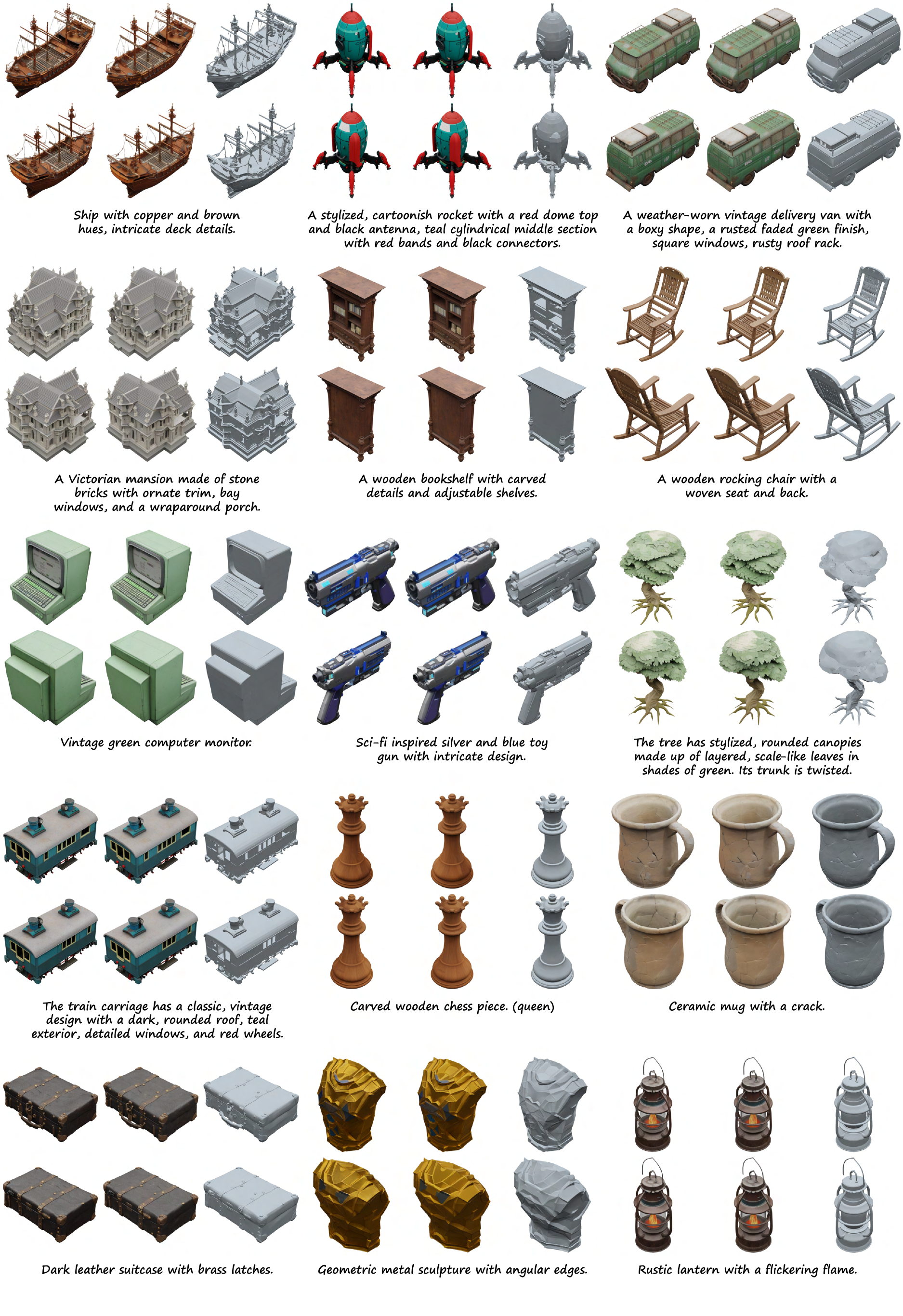}
    \caption{More results generated by \textsc{Trellis} with AI-generated text prompts. (From left to right: GS,  RF, and meshes)}
    \label{fig:more_assets_text}
\end{figure*}

\begin{figure*}[p]
    \centering
    \includegraphics[width=0.83\linewidth]{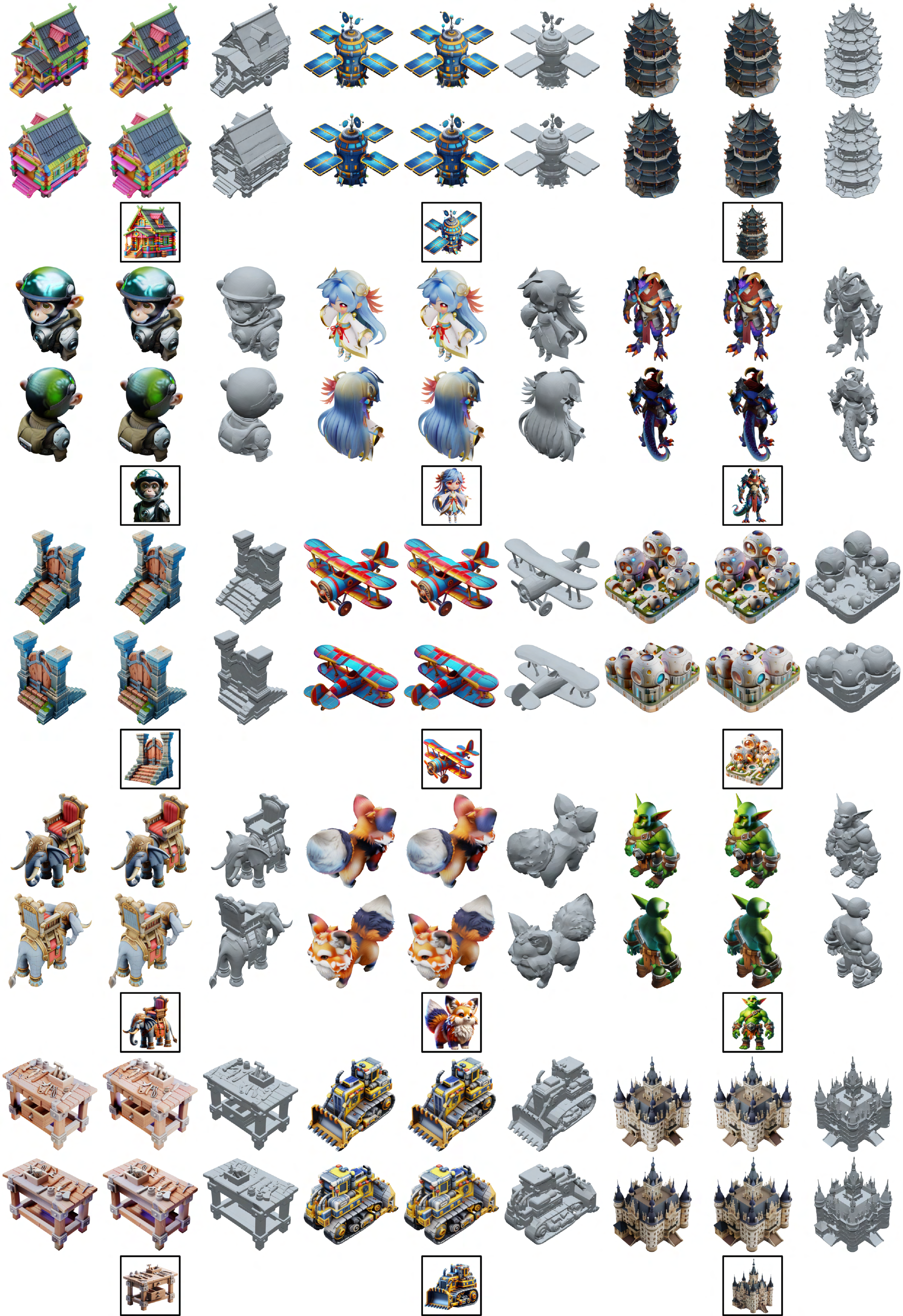}
    \caption{More results generated by \textsc{Trellis} with AI-generated image prompts. (From left to right: GS, RF, and meshes)}
    \label{fig:more_assets_image_aigen}
\end{figure*}

\begin{figure*}[p]
    \centering
    \includegraphics[width=0.83\linewidth]{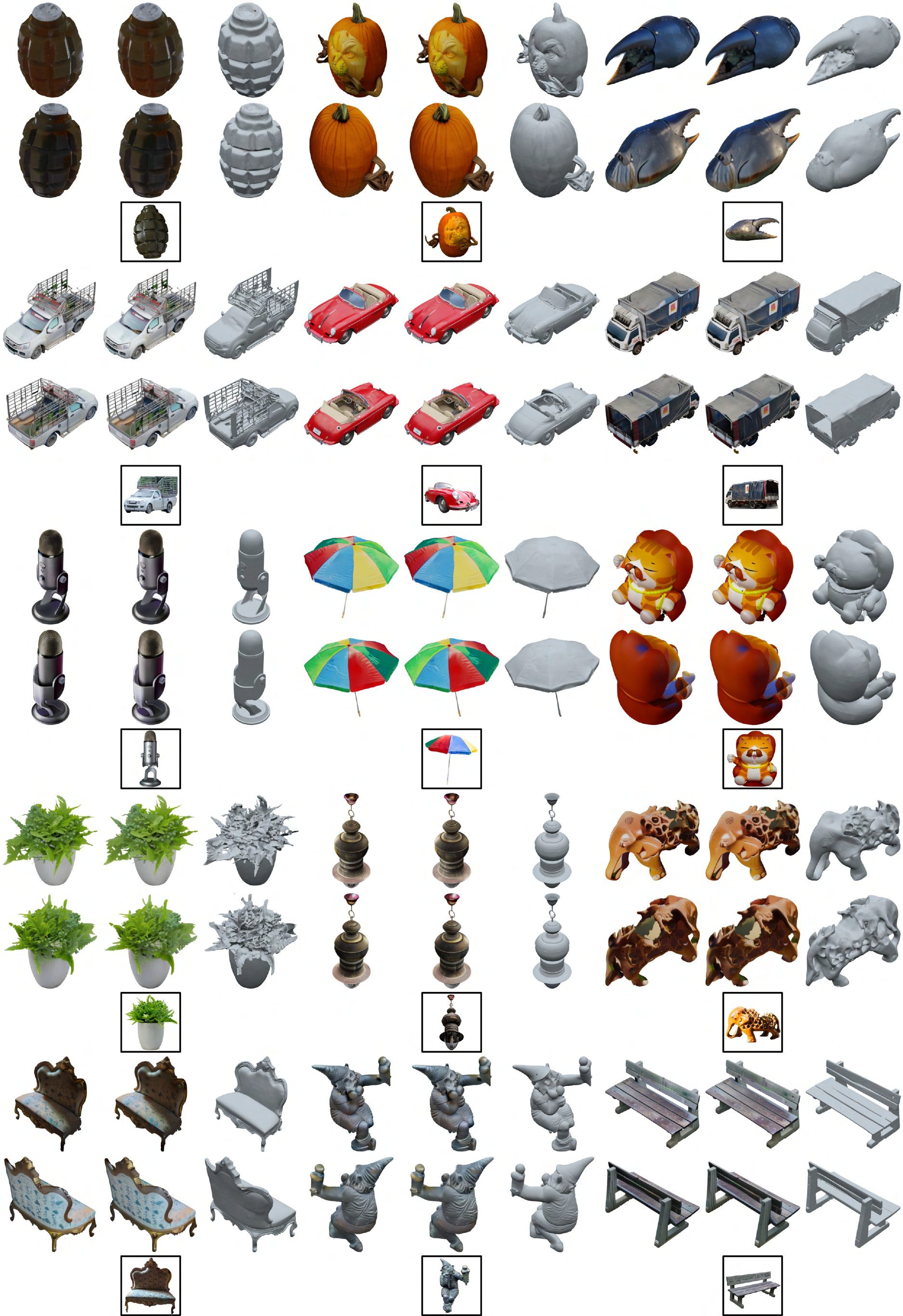}
    \caption{More results generated by \textsc{Trellis} with real-world image prompts from SA-1B. (From left to right: GS, RF, and meshes)}
    \label{fig:more_assets_image_real}
\end{figure*}

We present additional examples of 3D assets generated by our method. These include more text-to-3D results with AI-generated prompts in Fig.~\ref{fig:more_assets_text} and more image-to-3D results from both AI-generated images (Fig.~\ref{fig:more_assets_image_aigen}) and real world images (Fig.~\ref{fig:more_assets_image_real}). For real-world images, we use segmented objects from SA-1B~\cite{kirillov2023segment}, which feature challenging materials, geometries, and camera views.
Each $2\times3$ grid shows one generated asset, with front-left and back-right views in the top and bottom rows. Rendered images with 3D Gaussians (GS), Radiance Fields (RF), and meshes are displayed from left to right.

\subsection{More Comparisons}

\begin{figure*}[p]
    \centering
    \vspace{16pt}
    \includegraphics[width=\linewidth]{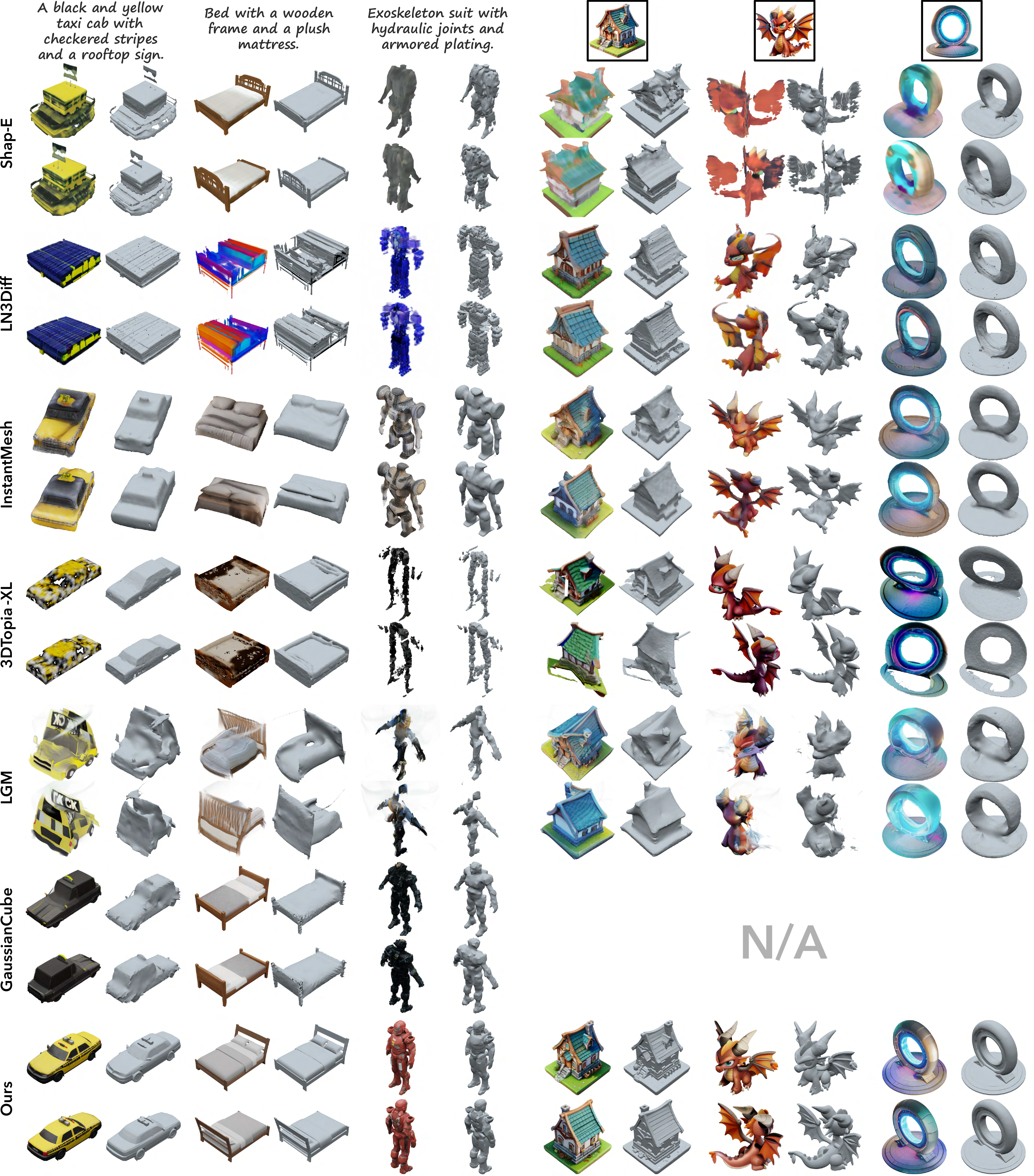}
    \caption{More comparisons of generated 3D assets by our method and prior works, with AI-generated text and image prompts.}
    \label{fig:more_comparisons}
    \vspace{16pt}
\end{figure*}

In Fig.~\ref{fig:more_comparisons}, we provide additional comparisons of 3D assets generated by our method and those produced by alternative approaches described in Sec.~\ref{sec:gen_results} in the main paper. 

\begin{figure*}[p]
    \centering
    \includegraphics[width=0.7\linewidth]{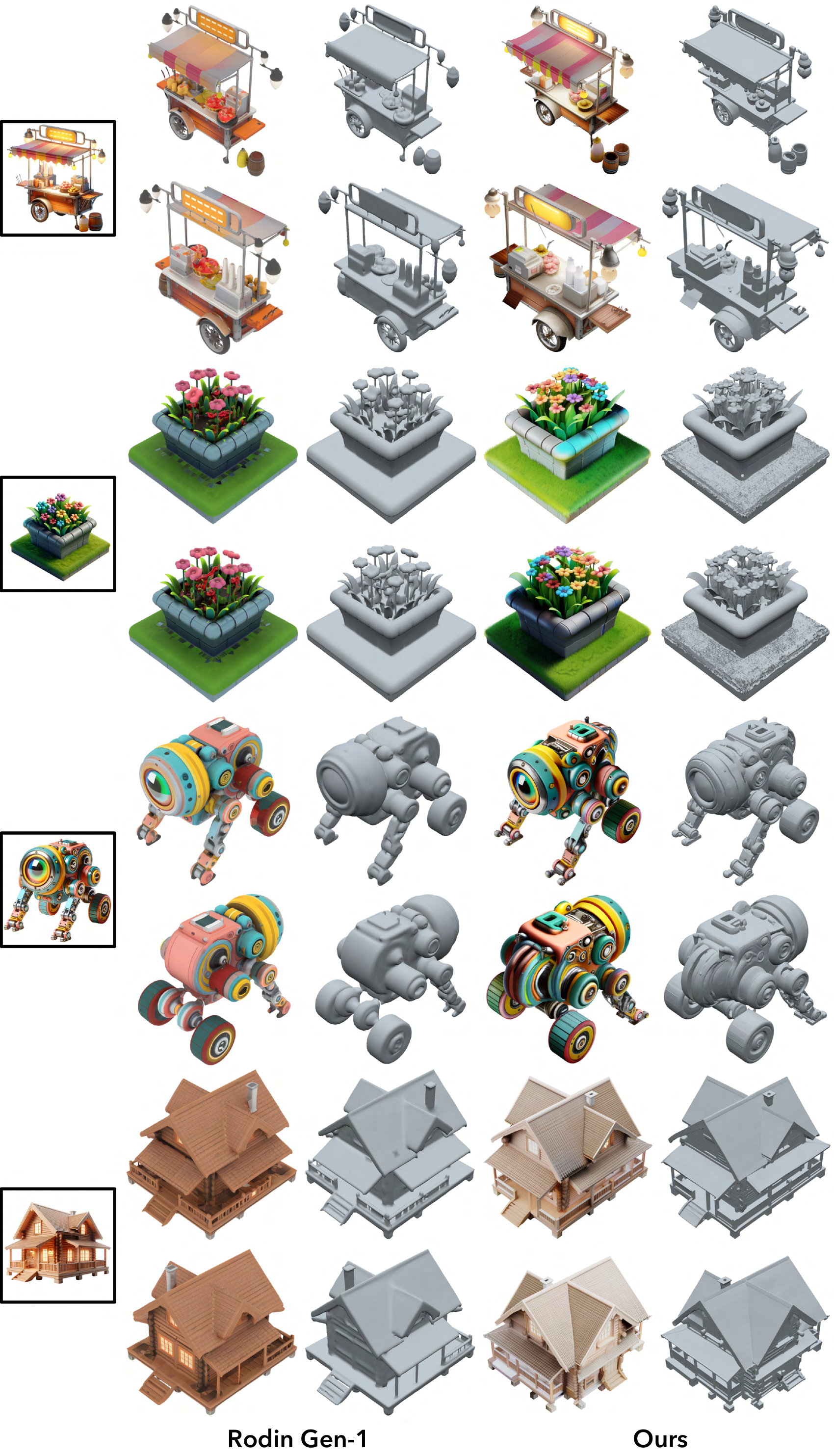}
    \caption{Comparisons between our method and a commercial-level 3D generation model, Rodin Gen-1 (with its default image-to-3D setting). Image prompts are generated by DALL-E 3. Our method exhibits more detailed geometry structures, while being trained solely on open-source datasets without commercial-specific designs.}
    \label{fig:more_comparisons_rodin}
\end{figure*}

Figure~\ref{fig:more_comparisons_rodin} further compares our method with the commercial-level 3D generation model, Rodin Gen-1\footnote{\href{https://hyperhuman.deemos.com/rodin}{https://hyperhuman.deemos.com/rodin}}, using its default image-to-3D generation setting. Our method exhibits more detailed geometry structures on these complex cases, while being trained solely on open-source datasets and without commercial-specific designs.

\subsection{3D Editing}
\begin{figure*}[p]
    \centering
    
    \vspace{16pt}
    \includegraphics[width=\linewidth]{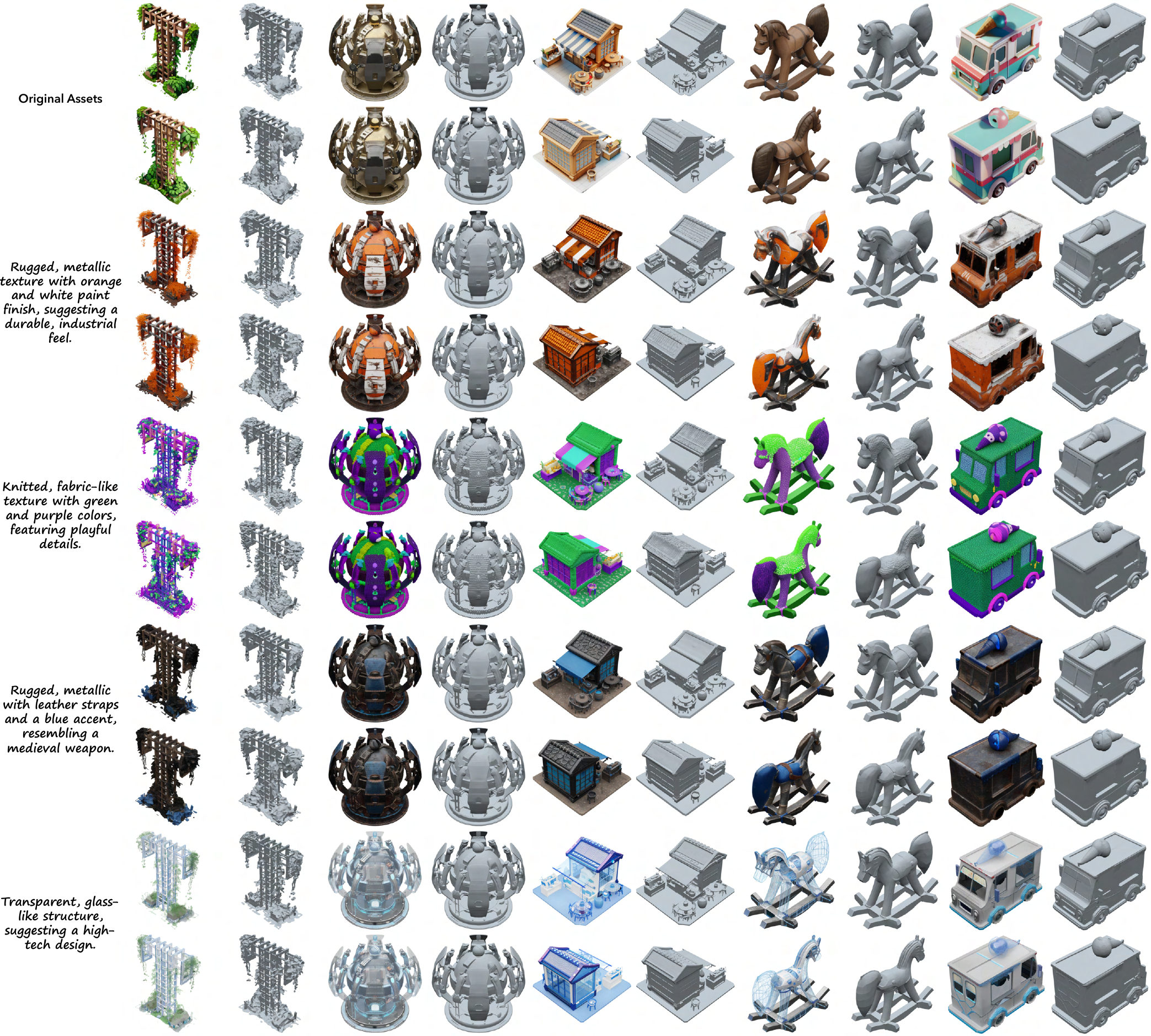}
    \caption{More examples of asset variations using \textsc{Trellis}. (Left: GS; Right: meshes)}
    \label{fig:more_editing_variations}
\end{figure*}

\begin{figure*}[p]
    \centering
    \includegraphics[width=\linewidth]{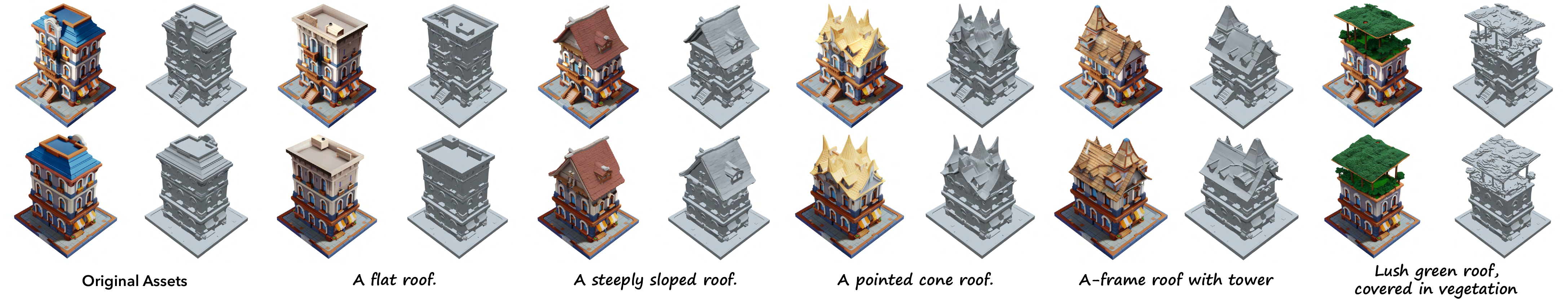}
    \caption{More examples of local editing, replacing the roof of the given building asset.}
    \label{fig:more_editing_local}
\end{figure*}

Figure~\ref{fig:more_editing_variations} and \ref{fig:more_editing_local} present additional editing results, highlighting the flexible capabilities of our method to edit and manipulate 3D assets.

\subsection{3D Scene Composition}
\begin{figure*}[p]
    \centering
    \includegraphics[width=0.83\linewidth]{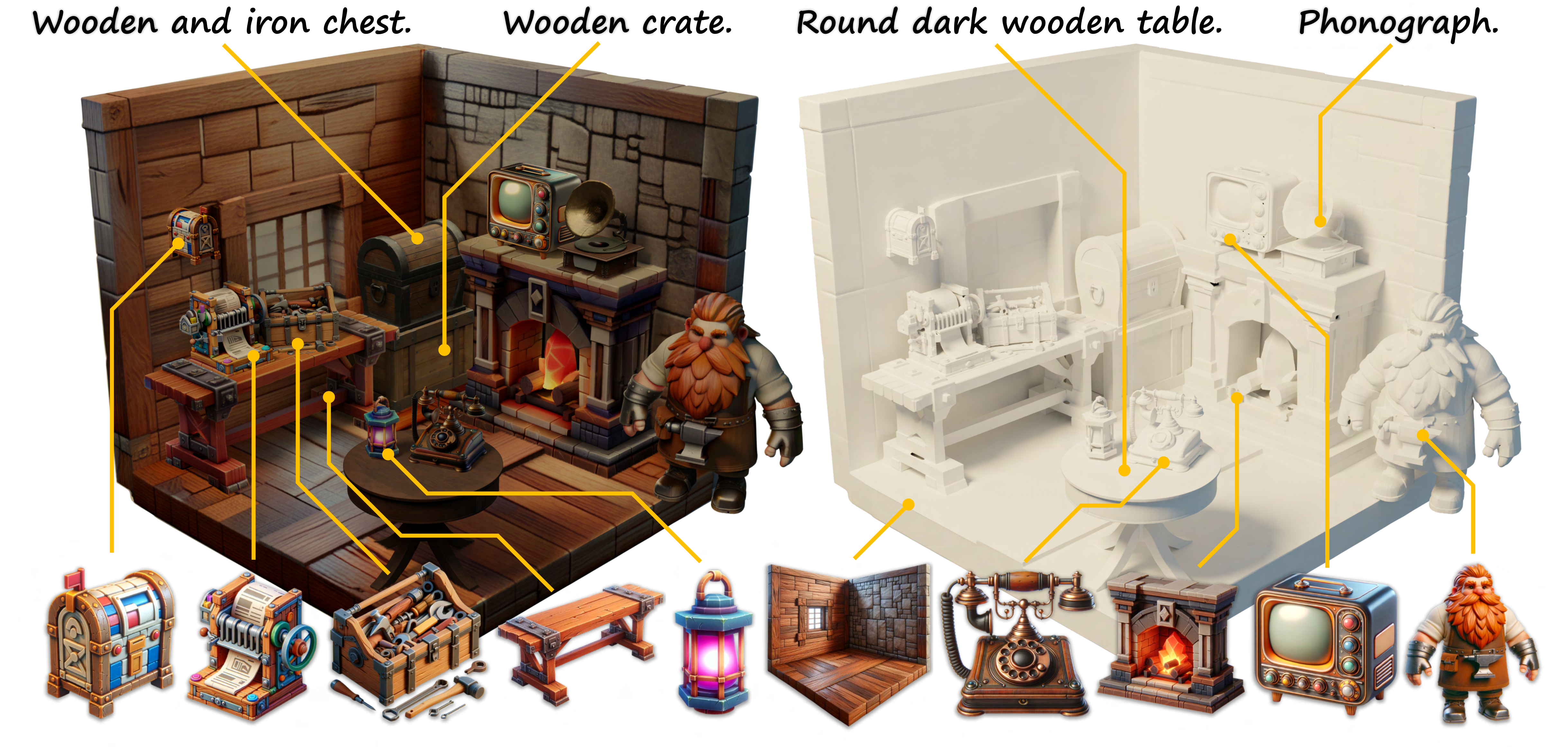}
    \vspace{-12pt}
    \caption{A dwarf blacksmith shop constructed with assets generated by \textsc{Trellis}. (\emph{Text and image prompts are linked with yellow lines})}
    \label{fig:scene_blacksmith}
\end{figure*}

\begin{figure*}[p]
    \centering
    \includegraphics[width=0.83\linewidth]{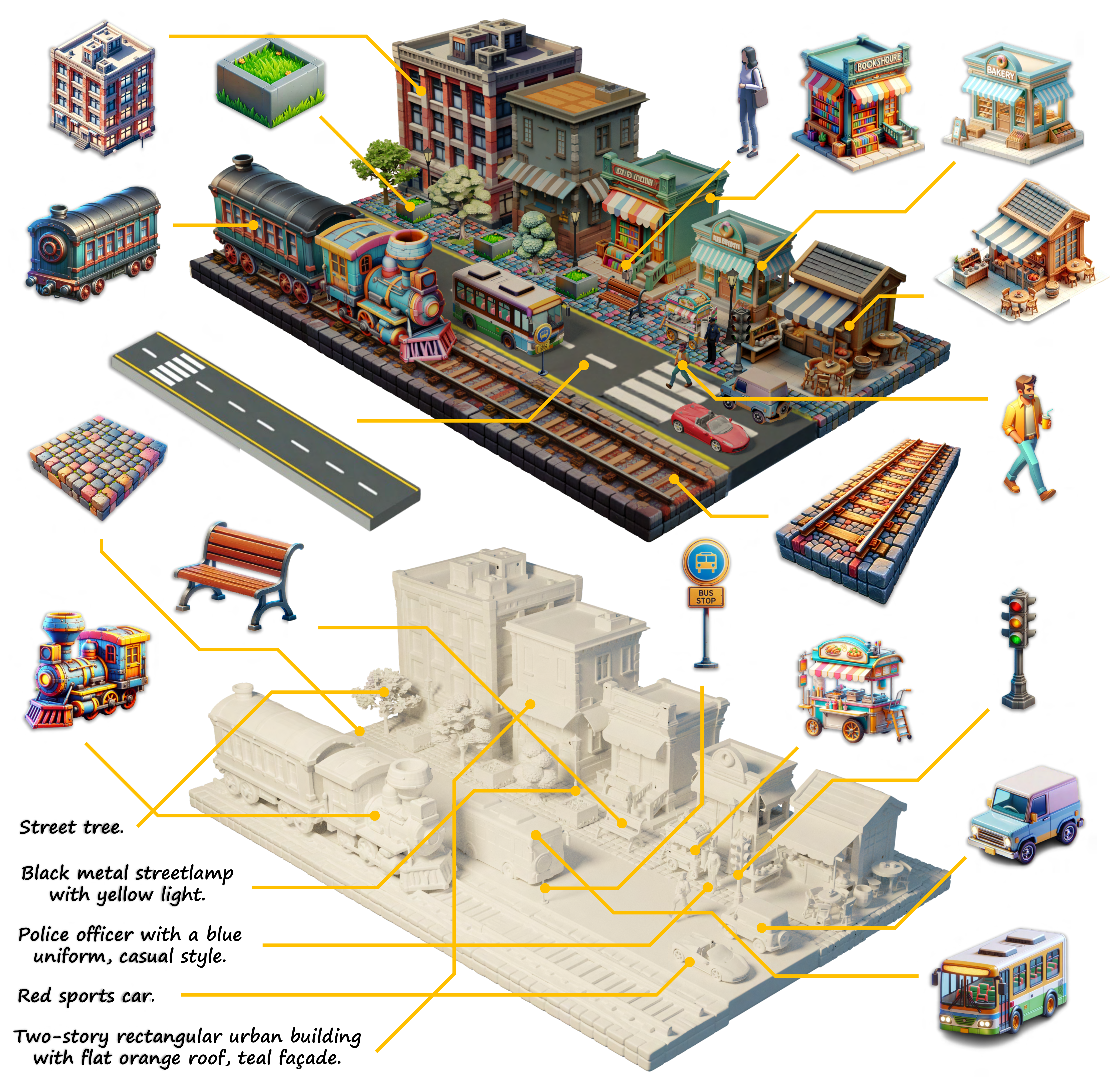}
    \vspace{-8pt}
    \caption{A vibrant streetview constructed with assets generated by \textsc{Trellis}. (\emph{Text and image prompts are linked with yellow lines})}
    \label{fig:scene_streetview}
\end{figure*}

Figure~\ref{fig:scene_blacksmith} and \ref{fig:scene_streetview} provide two supplementary visualizations of complex scenes constructed with assets from our model, demonstrating its potential for production use.

\section{Limitations and Future works}
While our model demonstrates strong performance on 3D generation, it still has some limitations. First, it uses a two-stage generation pipeline for the structured latent representation, which first generates the sparse structures, followed by the local latents on them. This approach can be less efficient than end-to-end methods that create complete 3D assets in a single stage. 

Second, our image-to-3D model does not separate lighting effects in the generated 3D assets, resulting in baked-in shading and highlights from the reference image. A potential improvement is to apply more robust lighting augmentation for image prompts during training and enforce the model to predict materials for Physically Based Rendering (PBR), which we leave for future exploration.

\end{document}